%% file: main.tex
\documentclass[11pt,letterpaper]{cmu}

\input{preamble}
\input{sections/abstract}

\begin{document}

\let\originalabscontent\abscontent
\renewcommand{\abscontent}{\input{sections/primer}\originalabscontent}
\maketitle
\let\abscontent\originalabscontent
\input{sections/introduction}

\input{sections/related}
\input{sections/problem}
\input{sections/methods}

\input{sections/experiments}

\input{sections/conclusion}

\bibliography{main}

\clearpage
\appendix
\input{sections/appendix}

\end{document}

%% file: preamble.tex
\usepackage[all]{hypcap}
\usepackage[comma,numbers,sort,compress]{natbib}
\usepackage{hyperref}[citecolor=magenta]

\hypersetup{
    colorlinks = true,
    citecolor = {magenta},
}

\usepackage{microtype}
\usepackage{graphicx}
\usepackage{subcaption}
\usepackage{booktabs}
\usepackage{float}

\usepackage{amsmath}
\usepackage{amssymb}
\usepackage{mathtools}
\usepackage{amsthm}
\usepackage{mathrsfs}
\usepackage{nicefrac}
\usepackage{dsfont}
\usepackage{enumitem}
\usepackage{float}

\usepackage{xspace}
\usepackage[capitalize,noabbrev]{cleveref}
\usepackage{subcaption}
\usepackage{wrapfig}
\usepackage{lipsum}
\usepackage{listings}

\usepackage{amsmath}
\usepackage{amssymb}
\usepackage{mathtools}
\usepackage{amsthm}
\usepackage{bbm}
\usepackage{tabularx}
\usepackage{multirow}
\usepackage[export]{adjustbox}
\usepackage{colortbl}
\usepackage{etoc}
\usepackage{algpseudocode}
\usepackage{setspace}

\usepackage{color}
\definecolor{deepblue}{rgb}{0,0,0.5}
\definecolor{deepred}{rgb}{0.6,0,0}
\definecolor{deepgreen}{rgb}{0,0.5,0}
\definecolor{boost_correct_to_correct}{HTML}{66C2A5}
\definecolor{default_correct_to_correct}{HTML}{fc8d62}
\definecolor{dup_correct_to_correct}{HTML}{8da0cb}
\definecolor{new_correct_to_correct}{HTML}{e78ac3}

\newcommand\pythonstyle{\lstset{
basicstyle=\ttfamily\footnotesize,
language=Python,
morekeywords={self, clip, exp, mse_loss, uniform_sample, concatenate, logsumexp},              
keywordstyle=\color{deepblue},
emph={MyClass,__init__},          
emphstyle=\color{deepred},   
stringstyle=\color{deepgreen},
frame=single,                       
showstringspaces=false
}}

\lstnewenvironment{python}[1][]
{
\pythonstyle
\lstset{#1}
}
{}

\newcommand\pythoninline[1]{{\pythonstyle\lstinline!#1!}}

\input{macro}

\makeatletter
\def\mathcolor#1#{\@mathcolor{#1}}
\def\@mathcolor#1#2#3{%
  \protect\leavevmode
  \begingroup
    \color#1{#2}#3%
  \endgroup
}
\makeatother

\usepackage[textsize=tiny]{todonotes}
\usepackage{algorithm}
\usepackage{amssymb}

\definecolor{myblue}{RGB}{224, 240, 255} 
\usepackage[capitalize,noabbrev]{cleveref}

\usepackage[textsize=tiny]{todonotes}

\usepackage[skins,theorems]{tcolorbox}
\Crefformat{equation}{#2Eq.\;(#1)#3}

\Crefformat{figure}{#2Figure #1#3}
\Crefformat{assumption}{#2Assumption #1#3}
\Crefname{assumption}{Assumption}{Assumptions}

\usepackage{crossreftools}
\usepackage[suppress]{color-edits}

\usepackage{dsfont}
\usepackage{nicefrac}

\definecolor{takeawaybg}{RGB}{244,246,228}
\newtcolorbox{analysisbox}[1][Takeaway]{
  enhanced,
  colback=takeawaybg, colframe=black,
  boxrule=1pt, arc=3pt,
  left=6pt, right=6pt, top=4pt, bottom=4pt,
  title={#1}, coltitle=white, fonttitle=\normalfont,
  attach boxed title to top left={xshift=4mm, yshift*=-\tcboxedtitleheight/2},
  boxed title style={colback=black, colframe=black, sharp corners, boxrule=0pt,
    left=4pt, right=4pt, top=2pt, bottom=2pt},
  before upper={
  \setlist[itemize]{
    leftmargin=12pt,
    labelsep=5pt,
    itemsep=0pt,
    parsep=0pt,
    topsep=0pt,
    partopsep=0pt
  }
},
}

\tcbset{
  aibox/.style={
    width=\linewidth,
    top=8pt,
    bottom=4pt,
    colback=blue!6!white,
    colframe=black,
    colbacktitle=black,
    enhanced,
    center,
    attach boxed title to top left={yshift=-0.1in,xshift=0.15in},
    boxed title style={boxrule=0pt,colframe=white,},
  }
}
\newtcolorbox{AIbox}[2][]{aibox,title=#2,#1}
\definecolor{lightblue}{rgb}{0.22,0.45,0.70}

\usepackage{pifont}
\usepackage{soul}
\definecolor{highlightmistake}{RGB}{255, 179, 179}
\definecolor{highlightcorrect}{RGB}{179, 255, 179}

\title{\textit{i}WAM: Efficient Action Prediction from World Models without Explicit Future Prediction}

\reportnumber{} 
\author[1]{Vansh Kapoor}
\author[1]{Max Sobol Mark}
\author[1]{Sreyas Venkataraman}
\author[1]{Max Simchowitz}
\author[1]{Aviral Kumar}

\affil[1]{Carnegie Mellon University}

\correspondingauthor={vanshk@cs.cmu.edu}
\projectwebsite={https://vansh28kapoor.github.io/iwam.github.io/}

\definecolor{lossGFT}{HTML}{247B78}
\definecolor{lossPredict}{HTML}{6553A6}
\definecolor{lossTarget}{HTML}{AF652F}

\definecolor{lossRoundTrip}{HTML}{9B466A}

%% file: macro.tex
\newcommand{\ignore}[1]{}

\newcommand{\methodname}{\textbf{\textit{i}WAM}}

\definecolor{blanchedalmond}{rgb}{1.0, 0.92, 0.8}
\definecolor{carmine}{rgb}{0.59, 0.0, 0.09}
\definecolor{lightblue}{rgb}{0.22,0.45,0.70}

\newtheorem{theorem}{Theorem}[section]

\newtheorem{proposition}[theorem]{Proposition}

\renewcommand{\mathbf}{\boldsymbol}

\makeatletter
\def\Ddots{\mathinner{\mkern1mu\raise\p@
\vbox{\kern7\p@\hbox{.}}\mkern2mu
\raise4\p@\hbox{.}\mkern2mu\raise7\p@\hbox{.}\mkern1mu}}
\makeatother

\numberwithin{equation}{section}

\definecolor{amaranth}{rgb}{0.9, 0.17, 0.31}
\definecolor{antiquebrass}{rgb}{0.8, 0.58, 0.46}
\definecolor{antiquefuchsia}{rgb}{0.57, 0.36, 0.51}
\definecolor{chromeyellow}{rgb}{0.31, 0.47, 0.26}

\usepackage[dvipsnames]{xcolor}
\definecolor{maj5}{HTML}{2b8cbe}
\definecolor{maj5Imp}{HTML}{084081}

\definecolor{seq5wo}{HTML}{d95f0e}
\definecolor{seq5woImp}{HTML}{662506}

\definecolor{seq5w}{HTML}{6a51a3}
\definecolor{seq5wImp}{HTML}{3f007d}

\definecolor{selfwo}{HTML}{d95f0e}
\definecolor{selfwoImp}{HTML}{662506}

\definecolor{selfw}{HTML}{6a51a3}
\definecolor{selfwImp}{HTML}{3f007d}

\definecolor{glorewo}{HTML}{d95f0e}
\definecolor{glorewoImp}{HTML}{662506}

\definecolor{glorew}{HTML}{6a51a3}
\definecolor{glorewImp}{HTML}{3f007d}

\definecolor{vstar}{HTML}{d95f0e}
\definecolor{vstarImp}{HTML}{662506}

%% file: sections/abstract.tex
\begin{abstract}
\textbf{Abstract:} World-action models (WAMs) build policies on pretrained video-generation models, on the premise that predicting how a scene evolves helps determine how to act. However, generating future scenes is expensive for real-time control. This motivates efficient proxies for future imagination, but how to train such proxies without sacrificing action-prediction quality remains unclear. An effective proxy need not generate a high-resolution frame; it only needs to provide enough information to distinguish between possible futures. This leads to the design principle behind \methodname{}: rather than denoising an entire future frame, can we condition the action policy on intermediate features from a single video-model evaluation? We show that the answer is yes. Exploiting the reversible dynamics of flow-matching models, we invert each training future into the corresponding latent noise vector. However, a single velocity evaluation at this latent does not necessarily yield informative features for action prediction. We therefore learn a bidirectional flow map that amortizes future-generation computation, enabling a single model evaluation to provide effective conditioning features. At inference, we sample a latent noise vector from the prior and condition the policy using just one video-backbone forward pass. Across simulated benchmarks and real robots, \methodname{} recovers the generalization benefits of WAMs while substantially reducing its inference cost.
\end{abstract}

%% file: sections/primer.tex

\noindent\begin{minipage}{\linewidth}
    \centering
    \raisebox{-0.5\height}{\includegraphics[width=0.63\linewidth]{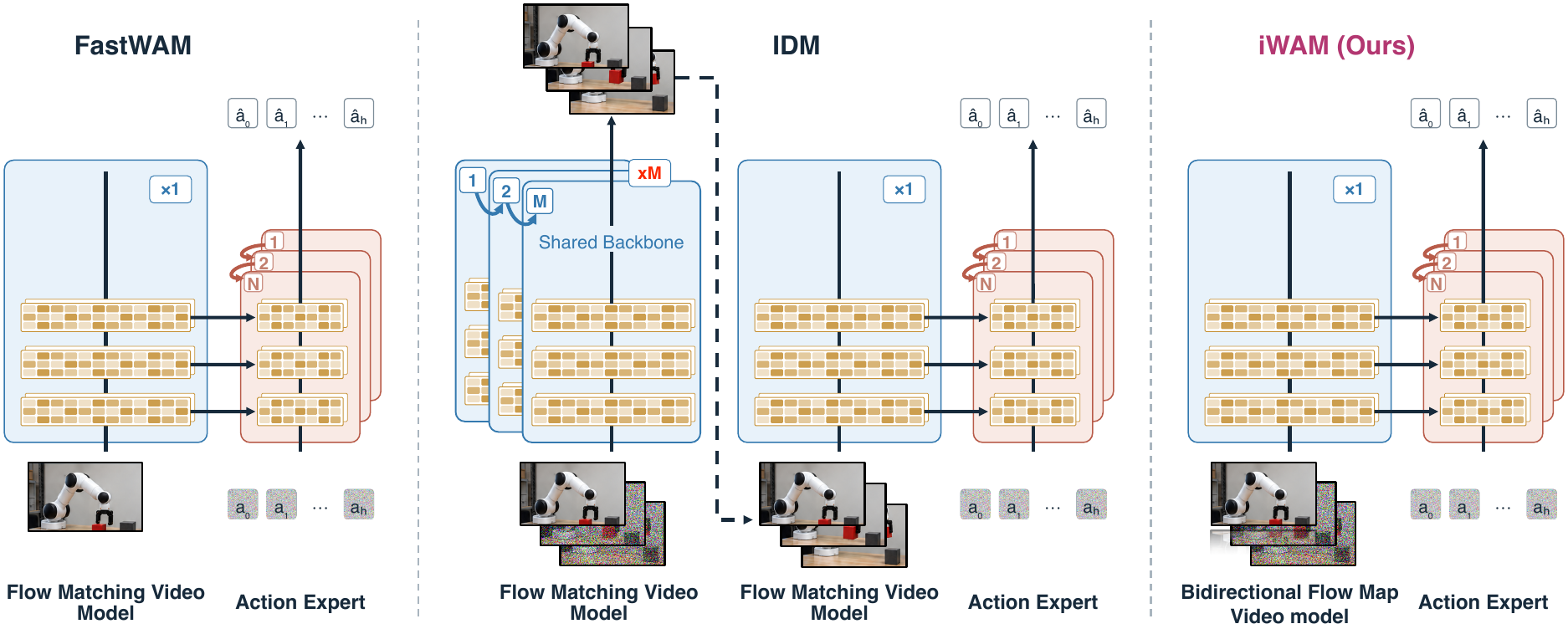}}\hfill
    \raisebox{\dimexpr-0.5\height-0.03\linewidth\relax}{\includegraphics[width=0.35\linewidth]{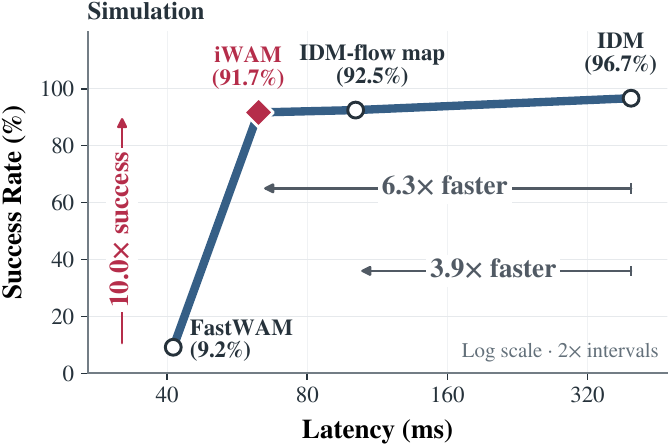}}\par
    \captionsetup{type=figure,skip=4pt}
    \caption{\textbf{Implicit World-Action Models (\methodname{}).} \textbf{Left (inference):} {\methodname{} provides future-specific conditioning within a single video-model evaluation, without explicitly generating a future.} It draws a noise endpoint from the flow prior and conditions the action expert on activations from a single noise-to-clean evaluation. \textbf{Right:} {DROID simulation success and inference latency. \methodname{} approaches IDM's success rate with approximately $6\times$ lower latency under a common two-step action-denoising budget. Success rates use each policy's evaluation budget (Section~\ref{sec:experiments}).}}
    \label{fig:primer}
    \label{fig:perf_latency}
    \label{fig:overview_inference}
\end{minipage}
\par\medskip

%% file: sections/introduction.tex
\vspace{-0.2cm}
\section{Introduction}
\label{sec:intro}
\vspace{-0.2cm}

Video generation models have become an attractive source of pretrained priors for robotic manipulation~\citep{wamsurvey,vid2world}. Trained on video data, which is far more abundant than robot data, they learn generalizable priors about objects, spatial relations, and physical dynamics that could help robotic policies generalize beyond limited robot demonstrations. World-action models (WAMs) adapt these models to predict actions by finetuning on robot data. Some jointly predict future video and actions~\citep{dreamzero,unifiedWM,motus}, while others generate future video to condition a separate action network, as in inverse dynamics models (IDMs)~\citep{unipi,causalWM}. During training, IDMs condition the action policy on activations from ground-truth futures; at deployment, these futures must instead be generated. Both approaches benefit from pretrained video priors but incur substantial inference cost from generating full, high-resolution videos, a major obstacle for real-time robotic control. Lighter-weight alternatives such as FastWAM~\citep{yuan2026fastwam} still train the video model to predict future observations alongside the action policy, but remove future observations from action conditioning entirely. Instead, the policy uses only intermediate features computed from the current observation, substantially reducing inference cost by avoiding future generation.

In principle, explicitly generating a future may be unnecessary if the video model's features already contain enough information about that future for action prediction. Is this possible? What information must these features retain, and how can we obtain it without generating the future itself? Typically, video generation spends \emph{test-time compute} to progressively transform a sample from the noise prior into a future frame. Our goal is to amortize this computation through training, so that a single forward pass produces features that support accurate action prediction. The challenge is therefore to preserve the useful information obtained through future generation while avoiding the full generation process.

Simply omitting that test-time computation does not guarantee that the remaining features contain the information needed to act. Our analysis shows that directly reading out actions from frozen video-backbone features using a FastWAM-style conditioning scheme generalizes poorly out of distribution. For example, when training on LIBERO this prior approach attains a strong in-distribution performance, but fails on perturbations in LIBERO-PRO (Sections~\ref{sec:liberopro} and~\ref{sec:experiments}). These results suggest that access to pretrained features alone is insufficient in this setting: the computation used to construct a possible future may also contribute to generalization. Our aim is to retain the benefits of this computation by learning a shortcut that provides informative future-conditioning features in a single forward pass.

To construct this shortcut, we exploit the reversible dynamics of flow-matching models. For a fixed conditioning vector, the underlying continuous flow defines a one-to-one correspondence between a noise vector and the future latent it generates. We can therefore invert any given future, the information an IDM would ordinarily condition on, to recover its corresponding noise. This raises a concrete question: can intermediate features from a single evaluation at this noise endpoint provide useful future conditioning without reconstructing the future itself? Reversibility alone does not guarantee this. A standard flow model predicts instantaneous velocities, and the corresponding future is obtained by integrating these velocities along the generative trajectory. Features from a single velocity evaluation need not capture the information that becomes accessible through this iterative generation process.

To make this correspondence useful, we propose \methodname{}, which learns a bidirectional flow map on robot data. Motivated by prior work on amortizing flow-matching processes~\citep{flowmaps,meanflow,shortcut}, this model approximates the average velocity of standard flow-matching over finite time intervals in both directions. This quantity summarizes the transformation along a trajectory segment, allowing the model to learn a shortcut to the computation required for future generation. We condition the action policy on intermediate features from a single evaluation of this bidirectional model. During training, we integrate each demonstrated future in the clean-to-noise direction to recover its noise endpoint, then extract features from one evaluation at that endpoint. This aligns the policy's conditioning features with the future corresponding to its supervising actions, and allows us to effectively fit the action expert. At inference, we sample an endpoint from the noise prior and extract features in the same way. The action policy combines these activations with features from past observations, requiring only one video-backbone evaluation for future conditioning.

Across simulated and real-world manipulation benchmarks~\citep{liberopro,droid,simevals}, \methodname{} recovers the generalization benefits of explicitly future-conditioned policies at the inference cost nearing that of approaches that discard future generation entirely. This clarifies what an action policy requires from a video model to generalize: neither a highly precise future nor observation-conditioned activations that leave the future unresolved, but sufficient information to identify the \emph{relevant mode of the future-video distribution}. Explicit future generation is sufficient to specify such a mode; the success of iWAM shows that it is not necessary.

%% file: sections/related.tex
\vspace{-0.2cm}
\section{Related Work}
\label{sec:related}
\vspace{-0.1cm}

\textbf{Video generation models for robot control.} Video models pretrained on large video datasets capture how objects and scenes evolve, providing a useful prior for robot control~\citep{wamsurvey,vid2world}. World-action models (WAMs) adapt this prior by finetuning on robot data and predicting actions from imagined futures~\citep{unipi,dreamitate,unifiedWM,motus,causalWM,dreamzero,cosmospolicy}. WAMs show promising generalization to unseen objects, scenes, and tasks, but repeatedly evaluating a large video backbone to generate futures makes inference slow. Our goal is to retain the benefits of conditioning on imagined futures without iterative generation.

\textbf{Efficient action prediction from video models.} Several methods avoid generating the future at inference. FastWAM~\citep{yuan2026fastwam} and UVA~\citep{uva} train the video model to predict the future alongside actions, but at test time predict actions from representations of the current observation alone. VPP~\citep{vpp} and VidMan~\citep{vidman} condition the action policy on intermediate features from a single forward pass of a video diffusion model, rather than on a generated video. We argue that such approaches are limited: their features are computed either from the current observation alone or from independently sampled noise, so during training they need not correspond to the future whose actions supervise the policy. In Section~\ref{sec:liberopro}, we show that removing the future from the conditioning hurts generalization to unseen environments, and in Section~\ref{sec:iwam}, we show that conditioning on independently sampled noise fits diverse data such as DROID poorly. \methodname{} also uses a single forward pass of the video backbone, but during training its activations correspond to the same future as the demonstrated actions.

\textbf{Few-step and bidirectional flow maps.} A large body of work reduces the cost of sampling from diffusion and flow models by learning to traverse the probability flow in a few large steps, including consistency models~\citep{consistency}, shortcut models~\citep{shortcut}, MeanFlow~\citep{meanflow}, terminal velocity matching~\citep{tvm}, and flow maps more generally~\citep{flowmaps}. The flow can also be run in reverse to recover the noise that generates a given sample: DDIM inversion~\citep{ddim} does so by integrating backward in many steps, and bidirectional consistency models~\citep{bcm} learn to do so in a few. These methods aim to generate or edit samples more cheaply. We use a bidirectional flow map for a different purpose, and never decode the future at all: during training, the clean-to-noise direction recovers the noise endpoint of the demonstrated future, and the noise-to-clean direction provides, in a single evaluation, the activations on which the action policy is conditioned.

%% file: sections/problem.tex
\vspace{-0.2cm}
\section{Preliminaries, Notation, and Problem Statement}
\vspace{-0.2cm}
\textbf{Flow-Matching.} 
Flow Matching~\citep{originflow} is a family of generative models that learn a time-dependent velocity field that transports a simple prior noise distribution, such as Gaussian noise, to the data distribution. Formally, given samples $z_1 \sim p_{\mathrm{prior}}$ and $z_0 \sim p_{\mathrm{data}}$, let $x_\tau= a_\tau z_0 + b_\tau z_1$ denote the interpolated state between the sampled data and noise endpoints at flow time $\tau\in[0,1]$, where $a_\tau$ and $b_\tau$ are the chosen time schedules satisfying $x_0=z_0$ and $x_1=z_1$. The corresponding conditional velocity~\citep{originflow} is
$u_\tau = x'_\tau = a'_\tau z_0 + b'_\tau z_1$,
where $'$ denotes the derivative with respect to flow time. 
For conditional generation with conditioning information $y$, a neural network parameterized by $\theta$ is trained to predict the marginal velocity field
$ v^\tau(x_\tau \mid y) =\mathbb{E}\left[u_\tau \mid x_\tau, y\right]$
using the flow-matching objective
\begin{align}
    \mathcal{L}_{\mathrm{FM}}(\theta)
    &=
    \mathbb{E}_{\tau,z_0,z_1,y}
    \left[
        \left\|
        v^\tau_\theta(x_\tau \mid y) - u_\tau
        \right\|_2^2
    \right].
\end{align}
At inference time, a sample $z_1 \sim p_{\mathrm{prior}}$ is drawn and transported along the learned flow trajectory by integrating the velocity field from the noise endpoint to the data endpoint, yielding a generated sample $z_0$. Here, $z_\tau$ denotes the state along the learned flow trajectory at flow time $\tau$, satisfying $\frac{d z_\tau}{d\tau}=v_\theta^\tau(z_\tau\mid y)$.

\textbf{Flow maps and shortcut models.} Flow map models~\citep{flowmaps} instead learn to directly transport a sample between two flow times $t,s \in [0,1]$. We parameterize the flow map through its average velocity $v_\theta^{t\rightarrow s}(z_t \mid y)$~\citep{meanflow}, such that
\begin{align}
    z_s
    &=
    z_t
    +
    (s-t)\,
    v_\theta^{t\rightarrow s}(z_t \mid y).
\end{align}
\begin{wraptable}{r}{0.49\textwidth}
    \caption{\textbf{Flow model and flow map notation.}}
    \vspace{-0.2cm}
    \label{tab:flow_terminology}
    \footnotesize
    \setlength{\tabcolsep}{3pt}
    \renewcommand{\arraystretch}{1.12}
    {
    \begin{tabularx}{\linewidth}{@{}l >{\raggedright\arraybackslash}X@{}}
    \toprule
    \textbf{Notation} & \textbf{Definition} \\
    \midrule
    $z_0,\ z_1$ & Clean and noise endpoints \\
    $t=0,\ t=1$ & Clean and noise flow times \\
    $x_\tau$ & Interpolated state at flow time $\tau$ \\
    $u_\tau$ & Conditional velocity $\frac{dx_\tau}{d\tau}$ at flow time $\tau$ \\
    $v_\theta^\tau(z \mid y)$ & Predicted marginal velocity at state $z$ and time $\tau$, conditioned on $y$ \\
    $z_\tau$ & State along the learned flow trajectory at time $\tau$ \\
    $v_\theta^{t\rightarrow s}(z_t \mid y)$ & Average velocity from flow time $t$ to $s$: $\frac{1}{s-t}\int_t^s v_\theta^\tau(z_\tau \mid y)\,d\tau$ \\
    \bottomrule
    \end{tabularx}}
    \vspace{-0.2cm}
\end{wraptable}
Unlike the marginal velocity in flow matching, the average velocity describes the displacement over the entire interval from $t$ to $s$. A flow map can therefore move directly between distant points along the flow without integrating the marginal velocity field over many intermediate steps. We consider flow maps defined over arbitrary time pairs $t,s\in[0,1]$, allowing both noise-to-clean ($s<t$) and clean-to-noise ($s>t$) transport, and refer to this as a \emph{bidirectional flow map}.
Flow maps can be trained with several self-distillation objectives~\citep{flowmaps}. Methods such as MeanFlow~\citep{meanflow} and TVM~\citep{tvm} achieve state-of-the-art one-step image generation, but require network derivatives, which are costly and can destabilize training at video scale. We instead use shortcut-style model training~\citep{shortcut}, which is related to~\citet{progressive} and bootstraps larger steps from compositions of smaller ones using only forward evaluations of the network.

\textbf{Problem formulation.} Given the current observation $o_0$ and language instruction $\ell$, let $c=(o_0,\ell)$ denote the conditioning context. A video model represents the conditional distribution of the ground truth future observations $o_{1:N}=(o_1,\ldots,o_N)$. We parameterize this distribution using a conditional flow model with marginal velocity field $v^\tau_{\theta_v}(z_\tau \mid c)$, where $z_0=o_{1:N}$ is the clean endpoint and $z_1\sim p_{\mathrm{prior}}$ is the noise endpoint. At inference, future observations are generated by sampling $z_1\sim p_{\mathrm{prior}}$ and integrating the flow from $\tau=1$ to $\tau=0$, yielding the predicted future observations $\hat{o}_{1:N}=z_0$. We denote the action policy by $\pi,$ which can either be a branch of a joint video-action prediction model or a separate downstream network, as in IDMs. We let $h_{\theta_v}(\cdot)$ denote the intermediate-layer features of the video model while computing $v^0_{\theta_v}$ on the given input. The various approaches below differ only in the inputs provided to $h_{\theta_v}.$

IDM conditions on features computed from the future observations and the action policy predicts:
\begin{align}
    a_{1:N} &\sim \pi\!\left(
    \cdot \mid
    h_{\theta_v}(c, \tilde{o}_{1:N})
    \right),
\qquad
    \tilde{o}_{1:N} =
    \begin{cases}
        o_{1:N}, & \text{during training},\\
        \hat{o}_{1:N}, & \text{at inference}.
    \end{cases}
\end{align}
In contrast, FastWAM-style approaches~\citep{yuan2026fastwam, gigaworld, uva} omit the future-observation tokens:
\begin{align}
    a_{1:N} &\sim \pi\!\left(
    \cdot \mid
    h_{\theta_v}(c)
    \right).
\end{align}
The key distinction therefore lies in the choice of $h_{\theta_v}$ utilized by the action policy. These two formulations highlight a fundamental trade-off. IDM directly ties action prediction to the generated future $\tilde{o}_{1:N}$, which is obtained by integrating the flow at inference time. This means that approaches based on IDM do spend additional \emph{test-time compute} to produce an action. From the literature on language models~\citep{snell2024scaling}, we know that spending more test-time compute results in better generalization. In contrast, FastWAM-style approaches completely avoid this test-time compute overhead by not explicitly conditioning on any specific future, but suffer from worse generalization properties, as we show below. In this paper, we will identify a choice of $h_{\theta_v}$ that ensures generalization to the level of IDMs, but with inference costs similar to FastWAM. 
Our goal is therefore twofold: first, to determine whether conditioning the action policy on the future is important for action policies to generalize; and second, to determine whether this conditioning can be provided without explicitly generating that future. We study the former question in Section~\ref{sec:liberopro}, and then use these findings to develop a more inference-efficient WAM.
\vspace{-0.2cm}

\section{Does Removing Future Information Hurt Generalization?}
\label{sec:liberopro}
\vspace{-0.1cm}

\noindent\begin{minipage}{\linewidth}
\begin{wrapfigure}{r}{0.53\textwidth}
    \centering
    \vspace{-0.5cm}
    \includegraphics[width=0.90\linewidth]{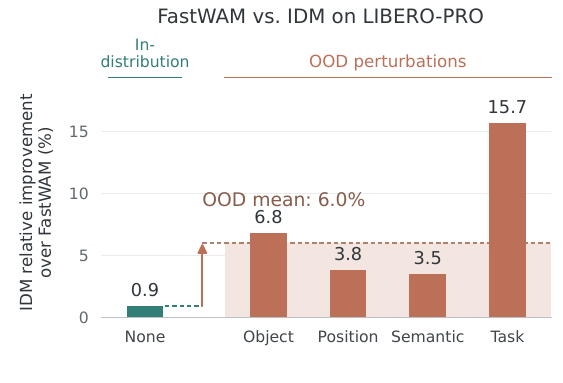}
    \vspace{-0.45cm}
    \caption{\textbf{FastWAM nearly matches IDM in-distribution but falls behind on OOD tasks and scenes.} With two-stage training on LIBERO-PRO, the relative gap grows from 0.9\% in-distribution to 6.0\% in mean OOD success.}
    \label{fig:idm_fastwam_advantage}
    \vspace{-0.3cm}
\end{wrapfigure}
IDMs explicitly condition the action policy on a predicted future at test time, which provides a strong signal for action prediction, but obtaining this future requires running an expensive iterative video generation process at inference time. FastWAM-style approaches avoid this cost by conditioning the action policy only on $h_{\theta_v}(c)$, and prior work has shown that this can perform well on standard in-distribution benchmarks~\citep{yuan2026fastwam}. This raises a natural question: \emph{when does explicitly conditioning on a specific future actually matter, and in particular, when does removing it begin to hurt performance?} We study this question by comparing the performance of policies conditioned on features obtained without spending any test-time compute at all (i.e., FastWAM-style) against those obtained by conditioning on the entire future imagination (i.e., IDM-style $h_{\theta_v}(c,\tilde{o}_{1:N})$).



\par

\end{minipage}

\begin{table}[t]
\centering
\caption{\textbf{FastWAM vs.\ IDM on DROID.} Success rates (\%) in tasks unseen during training, over 20 trials per task. Simulation results are averaged over replanning horizons of 16 and 24 steps; for Blocks $\rightarrow$ Bowl, success requires all three placements.}
\label{tab:fastwam_vs_idm}
\vspace{-0.2cm}
\small
\setlength{\tabcolsep}{3.5pt}
\renewcommand{\arraystretch}{1.0}
\begin{tabular}{@{}l cccc cccc@{}}
\toprule
& \multicolumn{4}{c}{Simulation} & \multicolumn{4}{c}{Real world} \\
\cmidrule(lr){2-5} \cmidrule(l){6-9}
Method
& Cube $\rightarrow$ Bowl & Can $\rightarrow$ Mug & Banana $\rightarrow$ Bin & Avg.
& Blocks $\rightarrow$ Bowl & Cup $\xrightarrow{\text{Pen}}$ Table & Fold Towel & Avg. \\
\midrule
FastWAM & 0.0 & 0.0 & 27.5 & 9.2 & 0.0 & 30.0 & 10.0 & 13.3 \\
IDM & \textbf{97.5} & \textbf{92.5} & \textbf{100.0} & \textbf{96.7} & \textbf{85.0} & \textbf{85.0} & \textbf{70.0} & \textbf{80.0} \\
\bottomrule
\end{tabular}
\end{table}
\textbf{Setup.} We compare FastWAM and IDM in two settings, training a separate set of models for each. In both, the two methods share the same Wan2.2-5B video backbone~\citep{wan2025} and 1B-parameter action expert. In our main setting, we train on DROID~\citep{droid}, a large and diverse dataset of real-world manipulation demonstrations, and evaluate the resulting policies in environments not seen during training: three pick-and-place tasks in the DROID simulation benchmark~\citep{simevals}, and three real-world tasks on a physical robot. In the second setting, we train on LIBERO~\citep{libero} and evaluate on LIBERO-PRO~\citep{liberopro}, which tests the policies under controlled perturbations to objects, initial states, and instructions. As such, the LIBERO benchmark alone is too saturated to be able to meaningfully differentiate the performance of various methods. Full training and evaluation details are provided in Section~\ref{sec:experiments}.











\begin{wraptable}{r}{0.5\linewidth}
\centering
\vspace{-0.3cm}
\caption{\textbf{FastWAM vs.\ IDM under LIBERO-PRO perturbations.} \emph{joint} denotes joint training of the video model and action policy, whereas \emph{disj} denotes the two-stage training procedure.}
\label{tab:liberopro_fastwam_idm}

\setlength{\tabcolsep}{5pt}
\renewcommand{\arraystretch}{1.}
\small

\begin{tabular}{lccccc}
\toprule
\textbf{Approach}
& \textbf{None}
& \textbf{Obj.}
& \textbf{Pos.}
& \textbf{Sem.}
& \textbf{Task} \\
\midrule

IDM (joint)
& \textbf{97.9} & \textbf{83.7} & \textbf{2.7} & \textbf{96.3} & \textbf{23.7} \\

FastWAM (joint)
& 96.9 & 78.0 & 0.9 & 83.7 & 15.3 \\

\addlinespace[2pt]
\midrule
\addlinespace[2pt]

IDM (disj)
& \textbf{98.5} & \textbf{84.9} & \textbf{10.9} & \textbf{98.0} & \textbf{23.6} \\

FastWAM (disj)
& 97.6 & 79.5 & 10.5 & 94.7 & 20.4 \\
\bottomrule
\end{tabular}
\end{wraptable}
\textbf{Findings.} On DROID, removing future conditioning substantially hurts performance. In simulation (Table~\ref{tab:fastwam_vs_idm}), FastWAM succeeds in 9.2\% of trials, compared to 96.7\% for IDM, and fails entirely on two of the three tasks. The gap persists on real robots: FastWAM averages 13.3\% success against 80.0\% for IDM. On Blocks $\rightarrow$ Bowl, FastWAM places at least one block in 80\% of trials but never completes all three placements, whereas IDM does so in 85\% of trials (per-horizon and per-placement results are reported in Tables~\ref{tab:replanning_results} and~\ref{tab:real_world_results}). Qualitative rollouts also reveal failures in FastWAM's instruction following: it manipulates objects unrelated to the instructed task or abandons the target object after a missed grasp (see Figure~\ref{fig:fastwam_failure_modes}). These failures occur despite the base video model's ability to generalize to these same unseen environments and accurately predict plausible futures, as also reflected in IDM's strong performance using these predictions. This gap suggests that FastWAM's action policy does not fully exploit the generalizable priors available in the pretrained video model, which IDM is able to utilize.


On LIBERO-PRO, the gap is smaller but consistent (Table~\ref{tab:liberopro_fastwam_idm}): averaged over the four perturbation types, IDM succeeds in 51.6\% of trials against 44.5\% for FastWAM with joint training, and 54.4\% against 51.3\% with two-stage training. Additional LIBERO-PRO results, including the effect of joint versus two-stage training, are provided in Appendix~\ref{appendix:liberopro}. {Figure~\ref{fig:idm_fastwam_advantage} shows that FastWAM falls further behind IDM as evaluation shifts to unseen scenes and tasks. The relative gap grows from 0.9\% in-distribution to 6.0\% in mean OOD success with two-stage training, and from 1.0\% to 16.0\% with joint training.} These results suggest that conditioning on features obtained from the current observation, as in FastWAM, may suffice when evaluation closely matches the training distribution, but becomes limiting on diverse data and in unseen environments, where features from an imagined future generalize better by drawing on the video model's large-scale pretraining. Next, we will build a simple mental model to understand this finding, which in turn will motivate a simple approach to address this issue.

\begin{analysisbox}[Takeaways: Only utilizing features of the current observation is insufficient.]
\begin{itemize}[itemsep=2pt]
\item FastWAM matches IDM in-distribution but falls well behind on unseen tasks and environments. 
\item The pretrained video prior generalizes to unseen settings, but FastWAM's observation-only conditioning fails to transfer it effectively to action prediction.
\end{itemize}
\end{analysisbox}

%% file: sections/methods.tex
\vspace{-0.2cm}
\section{\textit{i}WAM: Efficient WAMs Without Future Prediction}
\label{sec:iwam}
\vspace{-0.2cm}
The results above comparing FastWAM and IDM suggest a pattern similar to that observed in language models~\citep{snell2024scaling,setlur2025opt}: policies generalize better when we spend test-time compute generating future imaginations, as in IDM, but these gains disappear when we omit this process, as in FastWAM. In particular, not spending test-time compute appears most harmful in scenarios that demand generalization~\citep{setlur2025opt}, analogous to how a language model that only reasons with short chain-of-thought can have high in-distribution performance but may not generalize to slight perturbations of problems~\citep{gsmperturb,cotlength}. The natural next question is whether we can retain these generalization benefits without paying the full computational cost. At first glance, this may appear to require a ``free lunch'': if these benefits arise from spending substantial test-time compute generating a future, replacing this process with a single model evaluation might inevitably discard useful information. However, we also note that action prediction as such should not require reconstructing the full high-resolution future in principle, as long as we can extract a sufficiently informative proxy for it. This suggests a different possibility: we can \emph{amortize} enough of the generation process into a learned model to expose the information relevant for action prediction.

\vspace{-0.2cm}
\subsection{What Does Efficient Future Conditioning Require?}
\vspace{-0.1cm}
\begin{wrapfigure}{r}{0.55\textwidth}
    \centering
    \vspace{-0.3cm}
    \includegraphics[width=0.55\textwidth]{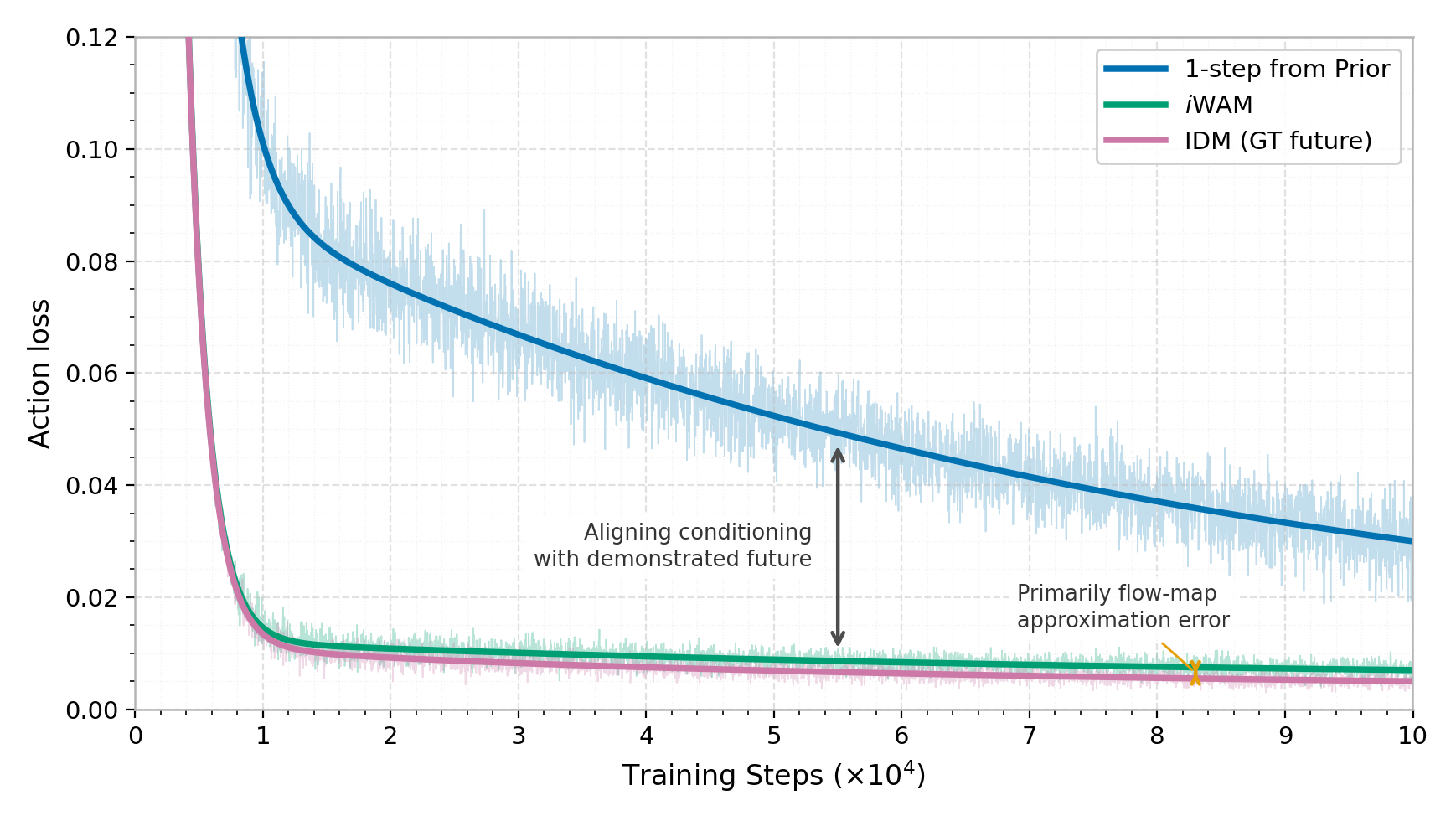}
    \vspace{-0.2cm}
    \caption{\textbf{Aligning the conditioning with the demonstrated future lowers action-training loss on DROID.}
    Sampling noise endpoints from the prior during training (1-Step from Prior) results in substantially higher training loss.
    \methodname{} aligns its predictive conditioning with the demonstrated future through the corresponding noise endpoint, largely closing the gap to IDM, which directly conditions on ground-truth futures during training. The small remaining gap may reflect approximation error in the learned bidirectional flow map.}
    \label{fig:ifc_training_loss}
    \vspace{-0.3cm}
\end{wrapfigure}
A natural starting point is to ask whether adding a small amount of test-time compute to FastWAM-style approaches can improve generalization. One straightforward option is to take one or a few integration steps along the video model's generative flow. At inference, running this process to completion yields exactly the generated future that IDM conditions on, which generalizes well in our experiments. This means that conditioning on features from a few integration steps could provide for a reasonable proxy for action prediction. Concretely, we consider conditioning the action expert on intermediate features from a \emph{single} video-backbone evaluation, instead of the many evaluations required to generate a future through standard flow matching. The simplest instantiation of this idea is as follows: for each training example $(c,o_{1},a_{1})$, sample a noise endpoint $z_1\sim p_{\mathrm{prior}}$ from the noise prior distribution of the flow-matching model, evaluate the velocity field once at $z_1$, and train the action expert to predict the demonstrated actions $a_{1}$ from features of the video model produced during this one forward pass to compute the velocity field. We refer to this variant as \emph{1-Step from Prior}. 
However, this approach fits the training data poorly: as shown in Figure~\ref{fig:ifc_training_loss}, it reaches a substantially higher action-training loss than IDM, even though both use the same amount of test-time compute during training. Why?

\begin{figure}[!t]
    \centering
    \vspace{-0.3cm}
    \includegraphics[width=0.95\linewidth]{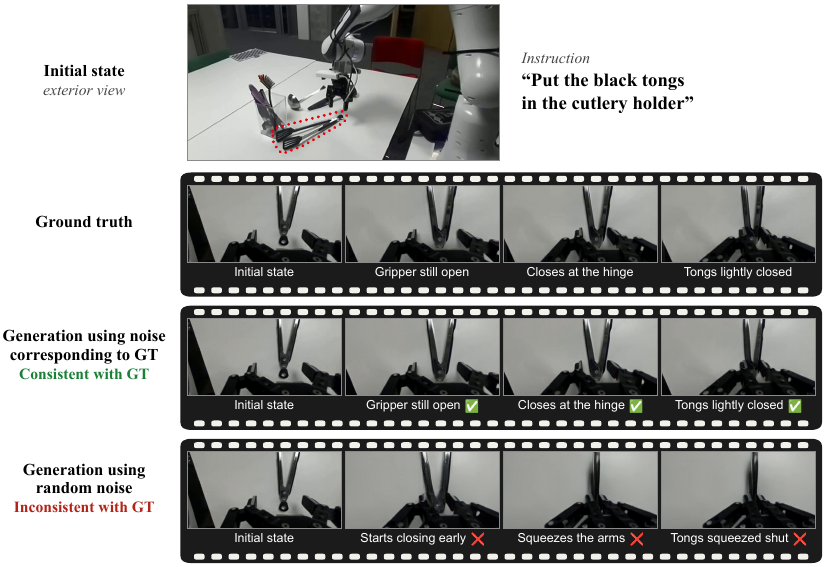}
    \vspace{-0.2cm}
    \caption{\textbf{Aligning the represented future with the demonstration.} Rows show the ground truth (top), the future generated from its corresponding noise endpoint (middle), and a future generated from random noise (bottom). Random-noise generation shows the gripper closing early around the tongs' arms and squeezing them shut; in the demonstration, it closes at the hinge and leaves the tongs only lightly closed. Generation from the corresponding noise endpoint preserves the demonstrated grasp timing and gripper configuration.}
    \label{fig:roundtrip_alignment}
    \vspace{-0.3cm}
\end{figure}
The difficulty is that a single evaluation of a standard flow model does not necessarily reveal features that uniquely direct towards the future imagination associated with the sampled noise endpoint. Under the full integrated flow, each $z_1$ deterministically maps to a particular clean future. Recovering that future, however, requires integrating the learned vector field over many steps. A standard flow-matching objective does not directly train a single evaluation to expose the future associated with a noise endpoint. In particular, the optimal one-step denoised estimate for standard flow-matching collapses to the conditional mean (see a theoretical argument in Proposition~\ref{proof:noise_endpoint_velocity}). Thus, the compute spent on this single evaluation starting from a noise endpoint cannot necessarily serve as useful test-time compute for action selection.

\begin{analysisbox}[Main Challenge: A Single Flow Evaluation Underutilizes Test-Time Compute]
A single flow-model evaluation consumes test-time compute without resolving a specific future, limiting its utility for action prediction.
\end{analysisbox}
A natural conclusion from the preceding argument is that amortizing the computation from multiple flow-model evaluations into a single evaluation could address this limitation. Our method in the next section will in fact build on this idea and learn ``shortcut'' models to amortize test-time compute. But is amortization alone sufficient? If the difficulty stems only from omitting the test-time compute and amortization will address it, then conditioning on a fully generated future should improve the fit in Figure~\ref{fig:ifc_training_loss}. We test this by sampling a noise endpoint $z_1$, fully integrating the flow, and conditioning the action policy on activations computed from the resulting future. During training, this differs from IDM only in using a sampled future in place of the actual future appearing in the data. Yet its training-loss curve remains close to that of the prior-noise variant and substantially above IDM's. Thus, even completing the full generation process does not resolve the fitting problem.

This gap points to a second requirement: to make effective use of amortized test-time compute, we also need \emph{alignment} between the conditioning future and the actions used for supervision. Each demonstration pairs a realized future $o_{1}$ with the actions $a_{1}$ that produce it, so the representation supplied to the action policy should correspond to that same future. An independently sampled $z_1$ need not satisfy this requirement: under the full flow, it may correspond to a different plausible future requiring different actions. The action policy would then receive conditioning that represents one future while being supervised with actions that realize another. To avoid this mismatch, we need the noise endpoint that the flow maps to the demonstrated future $o_{1}$. Fortunately, the integrated flow is invertible~\citep{flowmaps,originflow}, so we can recover this endpoint with a standard flow model by integrating its velocity field in the clean-to-noise direction. However, this inversion still requires multiple model evaluations. It therefore introduces substantial computational cost during training, even though it is not required at test time.

Figure~\ref{fig:roundtrip_alignment} illustrates this alignment requirement. Even with the same initial observation and instruction, a future generated from random noise can depict different behavior. Mapping the demonstrated future to noise and back instead yields a represented future more closely aligned with the demonstrated actions, motivating the use of a demonstration-specific noise endpoint during action training.

\begin{analysisbox}[Desiderata: Effective future conditioning requires specificity and alignment]
\begin{itemize}[itemsep=2pt,topsep=-2pt]
\item \textbf{Future specificity:} A single model evaluation must capture information about some particular future associated with a sampled noise endpoint.
\item \textbf{Future-action alignment:} During policy training, the represented future must correspond to the demonstrated actions to avoid conflicting supervision.
\end{itemize}
\end{analysisbox}

\vspace{-0.2cm}
\subsection{Key Idea: Amortizing Generation with Bidirectional Flow Map}
\vspace{-0.1cm}

We now introduce our method for addressing both challenges above. For efficient action prediction, we need to amortize noise-to-clean computation into a single model evaluation that provides information about the corresponding future. For efficient training, we also need to amortize clean-to-noise computation so that we can recover noise endpoints aligned with demonstrated futures. We accomplish both by learning a \textbf{\emph{bidirectional flow map}} that directly predicts transport between arbitrary flow times in either direction.

We parameterize the bidirectional flow map through an average velocity
$v_{\theta_v}^{t\rightarrow s}(z_t\mid c)$ that directly transports a state from flow time $t$ to flow time $s$:
\begin{align}
    z_s
    &=
    z_t
    +
    (s-t)\,
    v_{\theta_v}^{t\rightarrow s}(z_t\mid c).
    \label{eq:flowmap_velocity}
\end{align}
Unlike standard flow-map training, which samples time pairs only in the noise-to-clean direction~\citep{flowmaps,meanflow,shortcut}, we train over arbitrary time pairs $(t,s),$ covering both directions. Given a demonstrated future $z_0=o_{1:N}$, the clean-to-noise direction directly produces its corresponding noise endpoint, and the noise-to-clean direction predicts transport from this endpoint back toward the same demonstrated future:
\begin{align}
    z^{o_{1:N}}_1 &= o_{1:N} + v_{\theta_v}^{0\rightarrow 1}\!\left(o_{1:N}\mid c\right),
    \qquad
    o_{1:N} = z^{o_{1:N}}_1 - v_{\theta_v}^{1\rightarrow 0}\!\left(z^{o_{1:N}}_1\mid c\right).
    \label{eq:clean_to_noise_flowmap}
\end{align}
Thus, the recovered noise endpoint $z^{o_{1:N}}_1$ and the noise-to-clean prediction are directly tied to the actual future $o_{1:N}$. Our method will train such bidirectional maps to enable efficient action prediction.

\begin{figure}[!t]
    \centering
    \vspace{-0.3cm}
    \includegraphics[width=0.75\linewidth]{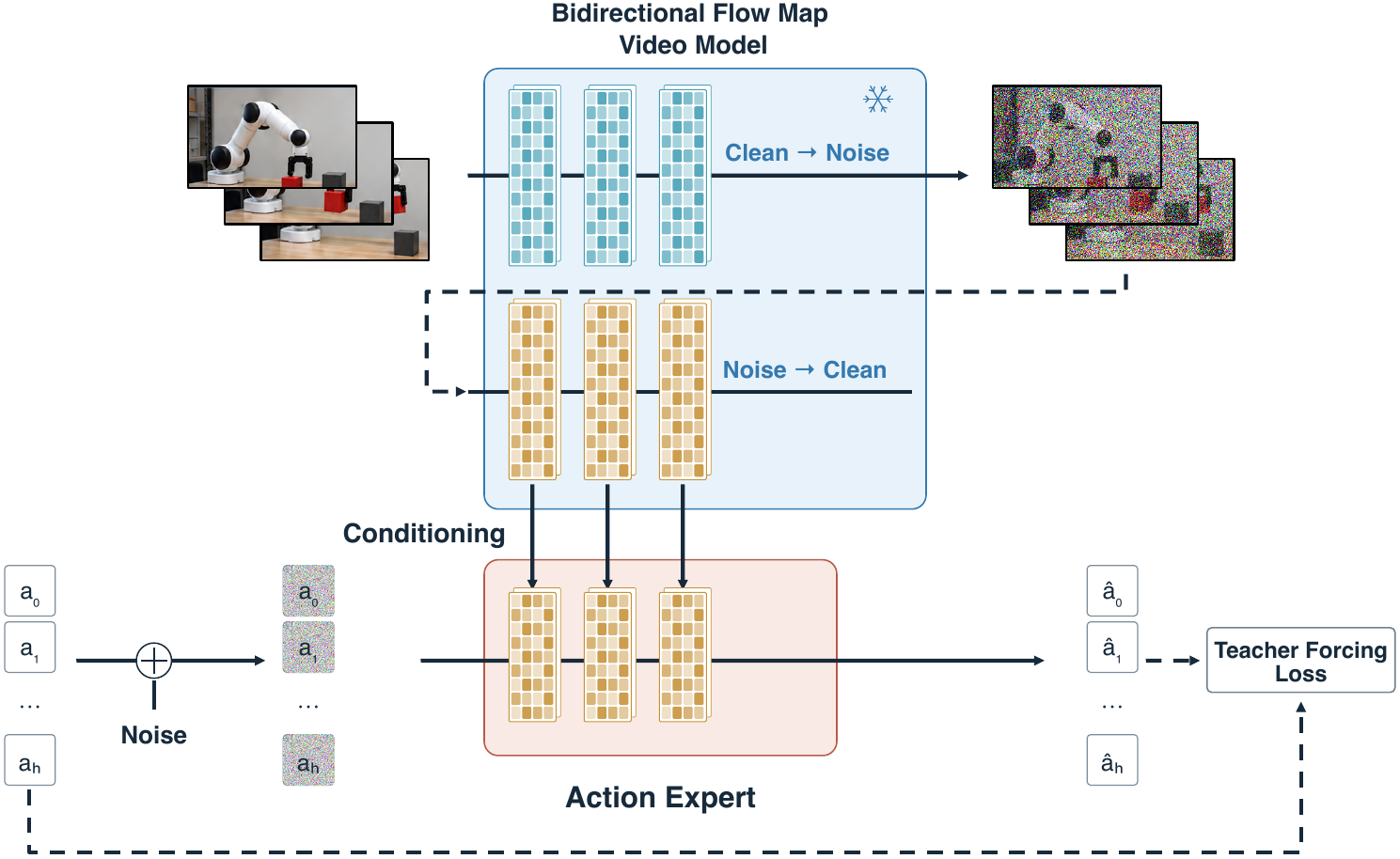}
    \caption{\textbf{Training \methodname{}.} The frozen flow map video model maps the demonstrated future to its noise endpoint with a single clean-to-noise evaluation, then performs a single noise-to-clean evaluation from that endpoint. The action expert is conditioned on the intermediate-layer activations of this second evaluation and trained to predict the demonstrated action chunk.}
    \label{fig:overview_training}
    \label{fig:overview}
\end{figure}

\vspace{-0.2cm}
\subsection{\methodname{}: Implicit World-Action Model}
\label{sec:implicit_wam}
\vspace{-0.1cm}

\begin{algorithm}[t]
\caption{Training and Inference for iWAM}
\label{alg:iwam}

\small
\noindent
\textbf{Require:}
Flow map model $\theta_v$, action model $\theta_a$

\vspace{0.45em}
\hrule
\vspace{0.45em}

\noindent
\begin{minipage}[t]{0.49\linewidth}
\textbf{Training}
\vspace{0.15em}

\begin{algorithmic}[1]
    \Repeat
        \State $(c,o_{1:N},a_{1:N}) \sim \mathcal{D}$

        \State $z^{o_{1:N}}_1
        \gets o_{1:N}
        + v_{\theta_v}^{0\rightarrow1}
        (o_{1:N}\mid c)$

        \State $h^{\mathrm{iWAM}}
        \gets h_{\theta_v}^{1\rightarrow0}
        (c,z^{o_{1:N}}_1)$

        \State $\theta_a
        \gets \operatorname{Update}\!\left(
            \theta_a,\,
            \mathcal{L}_{\mathrm{act}}
            (\theta_a;a_{1:N},h^{\mathrm{iWAM}})
        \right)$
    \Until{converged}
\end{algorithmic}
\end{minipage}
\hfill
\begin{minipage}[t]{0.47\linewidth}
\textbf{Inference}
\vspace{0.15em}

\begin{algorithmic}[1]
    \State $z_1 \sim p_{\mathrm{prior}}$

    \State $h^{\mathrm{iWAM}}
    \gets h_{\theta_v}^{1\rightarrow0}
    (c,z_1)$

    \State $\hat{a}_{1:N}
    \sim \pi_{\theta_a}
    (\cdot\mid h^{\mathrm{iWAM}})$

    \State \Return $\hat{a}_{1:N}$
\end{algorithmic}
\end{minipage}

\vspace{0.35em}
\hrule
\end{algorithm}

Our \methodname{} (\emph{implicit world-action model}) conditions the action policy directly on intermediate activations from the bidirectional video flow map. During training, given a demonstrated future $o_{1:N}$, we first evaluate the bidirectional flow map in the clean-to-noise direction to recover its corresponding noise endpoint $z^{o_{1:N}}_1$. We then evaluate the model once in the reverse, noise-to-clean direction at this endpoint, and use the intermediate-layer activations to condition the action policy. At inference, no demonstrated future is needed: we sample $z_1$ from the noise prior, perform a single noise-to-clean model evaluation, and condition the action policy on its intermediate activations. Thus, \methodname{} requires only one video-model evaluation at deployment.

Let $h_{\theta_v}^{t\rightarrow s}(c,z_t)$ denote the intermediate-layer activations used to predict
$v_{\theta_v}^{t\rightarrow s}(z_t\mid c)$.
Because
$h_{\theta_v}^{1\rightarrow0}(c,z^{o_{1:N}}_1)$
is used to predict transport from the recovered noise endpoint toward
$o_{1:N}$, these activations are trained to encode information about the demonstrated future without requiring that future to be explicitly generated. At inference, we instead sample $z_1\sim p_{\mathrm{prior}}$ and compute $h_{\theta_v}^{1\rightarrow0}(c,z_1)$ with a single video-model evaluation. The action policy then predicts actions as
\begin{align}
    a_{1:N} &\sim \pi\!\left(\cdot \mid h_{\theta_v}^{1\rightarrow0}(c,\tilde{z}_1)\right),
    \qquad
    \tilde{z}_1 =
    \begin{cases}
        z^{o_{1:N}}_1, & \text{during training},\\
        z_1, & \text{at inference},
    \end{cases}
    \label{eq:iwam_policy}
\end{align}
without explicitly generating the future associated with $z_1$. Algorithm~\ref{alg:iwam} summarizes the complete \methodname{} training and inference procedure. In this way, \methodname{} replaces future generation with a single video-model evaluation while retaining future-specific information that can guide action prediction.

\vspace{-0.3cm}
\subsection{Implementing \textit{i}WAM in Practice}
\vspace{-0.1cm}
\label{sec:iwam_practice}
To learn the flow map, we adopt shortcut-style training~\citep{shortcut}, which exploits the self-consistency property that a flow map over a given time interval is equivalent to the composition of flow maps over smaller subintervals. This enables bootstrapping predictions over longer intervals from shorter ones using only forward evaluations of the network. Intuitively, the bootstrap loss trains the model to approximate the result of two consecutive smaller transport steps with a single larger step, enabling transport between distant flow times in one model evaluation. Although shortcut-style objectives can yield lower one-step sample quality than derivative-based alternatives~\citep{meanflow,flowmaps}, they provide stable training when predicting many frames jointly, and \methodname{} requires features that can guide action prediction, rather than high-fidelity generations. Unlike the original shortcut formulation, we train over arbitrary time pairs $(t, s)$, enabling both the clean-to-noise and noise-to-clean transport. We train \methodname{} in two stages: 1) learning this bidirectional flow map and 2) learning an action expert on top of the flow map.

\textbf{Prompt guidance.}
Prompt guidance is important for video generation to generalize to unseen tasks and environments, and IDM-style approaches typically use classifier-free guidance (CFG). However, CFG combines two model evaluations to construct the guided field (Appendix~\ref{appendix:shortcut}); neither evaluation directly provides the guided intermediate activations that \methodname{} requires. Following Guidance-Free Training (GFT)~\citep{gft}, we instead condition the flow map on guidance strength $w$ and train it to predict the prompt-guided average velocity in a single evaluation. We use similar guidance strengths during action-policy training and inference to keep the guidance used for feature conditioning consistent.

\textbf{iWAM video model training.}
The video training objective in Equation~\eqref{eq:shortcut_gft} comprises three terms:
(i) the GFT loss $\mathcal{L}_{\mathrm{GFT}}$, which directly learns the marginal velocity field for arbitrary guidance strengths $w$;
(ii) shortcut self-consistency $\mathcal{L}_{\mathrm{SC}}$, which matches the predicted average velocity from $t$ to $s$ to a target computed from two half-steps of an exponential moving average (EMA) model; and
(iii) round-trip consistency $\mathcal{L}_{\mathrm{RT}}$, which enforces consistency between average velocities in opposite transport directions by matching the model's prediction from $s$ to $t$ to the EMA target from $t$ to $s$.
\begin{align}
\mathcal{L}_{\mathrm{video}}
&=\lambda_{\mathrm{FM}}\,
\mathcolor{lossGFT}{\underbrace{\mathbb{E}\!\Bigl[
\Bigl\|
\frac{1}{w}v_{\theta_v}^{\tau}(x_\tau\mid c,w)
+\left(1-\frac{1}{w}\right)\operatorname{sg}\!\left[v_{\theta_v}^{\tau}(x_\tau\mid\varnothing,1)\right]
-u_\tau\Bigr\|_2^2\Bigr]}_{\mathcal{L}_{\mathrm{GFT}}\,:\;\text{GFT loss}}}
\nonumber\\[5pt]
+ \lambda_{\mathrm{SC}} &\,
\mathcolor{lossPredict}{\underbrace{\mathbb{E}\!\Bigl[
\Bigl\|
 v_{\theta_v}^{t\rightarrow s}(x_t\mid c,w)
-\operatorname{sg}\!\left[\bar v^{t\rightarrow s} (x_t\mid c,w)\right]
\Bigr\|_2^2\Bigr]}_{\mathcal{L}_{\mathrm{SC}}\,:\;\text{Shortcut self-consistency loss}}} +\lambda_{\mathrm{RT}}\,
\mathcolor{lossRoundTrip}{\underbrace{\mathbb{E}\!\Bigl[
\Bigl\|
 v_{\theta_v}^{s\rightarrow t}(\tilde x_s\mid c,w)
-\operatorname{sg}\!\left[\bar v^{t\rightarrow s}(x_t\mid c,w)\right]
\Bigr\|_2^2\Bigr]}_{\mathcal{L}_{\mathrm{RT}}\,:\;\text{Round-trip consistency loss}}}.
\label{eq:shortcut_gft}
\end{align}
We sample $(o_{1:N},c)\sim\mathcal{D}$ and construct interpolated states $x_r$ between the demonstrated future and Gaussian noise, for $r\in[0,1]$. We set the guidance strength to $w=1/\beta$, where $\beta\sim\mathcal{U}(0,1)$. Time pairs $(t,s)$ are sampled equally from both transport directions: noise-to-clean ($s<t$) and clean-to-noise ($s>t$). For GFT, we drop the language instruction on 10\% of training examples while retaining the current observation $o_0$, denoting the resulting conditioning by $\varnothing$. Both these examples and the unconditional prediction in the GFT loss use $w=1$. The shortcut and round-trip consistency losses share a target computed using an exponential moving average (EMA) model with parameters $\bar\theta_v$. With midpoint $m=(t+s)/2$, this target is given by
\begin{align}
\tilde x_m
&=x_t+(m-t)\,v_{\bar\theta_v}^{t\rightarrow m}(x_t\mid c,w),
\nonumber\\
\bar v^{t\rightarrow s}(x_t\mid c,w)
&=\frac{1}{2}\left[
 v_{\bar\theta_v}^{t\rightarrow m}(x_t\mid c,w)
+v_{\bar\theta_v}^{m\rightarrow s}(\tilde x_m\mid c,w)
\right],
\nonumber\\
\tilde x_s&=x_t+(s-t)\,\operatorname{sg}\!\left[\bar v^{t\rightarrow s}(x_t\mid c,w)\right].
\label{eq:shortcut_target}
\end{align}
Both $\mathcal{L}_{\mathrm{SC}}$ and $\mathcal{L}_{\mathrm{RT}}$ use the same stop-gradient EMA target $\bar v^{t\rightarrow s}$, with $\operatorname{sg}$ denoting stop-gradient. While shortcut self-consistency matches the predicted prompt-guided average velocity to this target, it does not explicitly enforce consistency between the two transport directions. We therefore introduce $\mathcal{L}_{\mathrm{RT}}$ to encourage accurate round-trip transport, as motivated by prior work on bidirectional consistency models~\citep{bcm}. For efficiency, we reuse the EMA target and its endpoint $\tilde x_s$ to supervise the reverse transport without additional EMA evaluations. The weights $\lambda_{\mathrm{FM}}$ and $\lambda_{\mathrm{SC}}$ are determined by the batch allocation; the round-trip sampling fraction and weight schedule are provided in Appendix~\ref{appendix:droid_training}.

\textbf{\methodname{} Action Policies: Flow and MIP.} We train \methodname{} with both a flow-based action policy and the Minimal Iterative Policy (MIP) objective~\citep{mip}. MIP predicts actions in two denoising steps: an initial regression step followed by refinement. It nearly matches the performance of the flow-based policy, which uses multiple denoising steps at evaluation (Section~\ref{sec:experiments}).

%% file: sections/experiments.tex
\vspace{-0.4cm}
\section{Experiments}
\label{sec:experiments}
\vspace{-0.2cm}

The goal of our experiments is to evaluate whether \methodname{} retains the benefits of future conditioning observed in Section~\ref{sec:liberopro} with a single video-model evaluation. We study three questions:
\begin{samepage}
\begin{enumerate}[nosep]
    \item {Can \methodname{} retain performance that roughly matches IDM in environments not seen during training, without explicitly generating future imaginations?}
    \item {How does \methodname{} compare with IDM and FastWAM in task success and inference latency?}
    \item {When does training-time alignment with the demonstrated future matter most?}
\end{enumerate}
\end{samepage}

\textbf{Experimental setup.}
We evaluate on DROID~\citep{droid} and LIBERO-PRO~\citep{liberopro} to study generalization in diverse manipulation environments and under controlled distribution shifts.
DROID is a large-scale, real-world manipulation dataset containing 95K trajectories collected across 564 scenes using a Franka Panda arm with a Robotiq gripper. It captures similar manipulation behaviors across different objects, environments, and operators. We evaluate DROID-trained policies on three simulated pick-and-place tasks and three physical-robot tasks in environments not seen during training.
LIBERO-PRO evaluates policies trained on LIBERO~\citep{libero}, which comprises four 10-task suites covering spatial relationships, object variation, task goals, and long-horizon manipulation. It tests generalization through perturbations to object attributes, initial object positions, instruction phrasing, and task intent.
We train a separate set of policies on each dataset, using the same pretrained Wan2.2-5B video backbone and 1B-parameter action expert for all WAMs. DROID policies use the filtered training data described in Appendix~\ref{appendix:droid_training}; LIBERO policies are trained on the original demonstrations and evaluated on LIBERO-PRO without further finetuning. Full training and evaluation details are provided in Appendices~\ref{appendix:liberopro} and~\ref{appendix:droid_training}.

\begin{table*}[t]
\centering

\begin{minipage}[t]{0.49\textwidth}
\vspace{0pt}
\centering
\scriptsize
\setlength{\tabcolsep}{2.6pt}
\renewcommand{\arraystretch}{0.95}

\caption{\footnotesize
\textbf{Real-world performance on DROID.}
Success rates (\%) over 20 trials per task.
For Blocks $\rightarrow$ Bowl, columns report the fraction
of trials completing at least one, at least two, or all
three placements.}
\label{tab:real_world_results}

\resizebox{\linewidth}{!}{%
\begin{tabular}{l ccc c c c}
\toprule
& \multicolumn{3}{c}{Blocks $\rightarrow$ Bowl}
& Cup $\xrightarrow{\text{Pen}}$ Table
& Fold Towel
& \multirow{2}{*}{Avg.} \\
\cmidrule(lr){2-4}
\cmidrule(lr){5-5}
\cmidrule(lr){6-6}
Method
& $\geq 1$ & $\geq 2$ & 3
& Success & Success & \\
\midrule
FastWAM
& 80\% & 15\% & 0\%
& 30\% & 10\% & 13.3\% \\

IDM
& \textbf{100\%} & \textbf{90\%} & \textbf{85\%}
& 85\% & 70\% & 80.0\% \\

\midrule
\rowcolor{myblue}
\textit{i}WAM (MIP)
& \textbf{100\%} & \textbf{90\%} & 75\%
& \textbf{95\%} & \textbf{85\%} & \textbf{85.0\%} \\
\bottomrule
\end{tabular}%
}
\end{minipage}%
\hfill
~\vline~
\begin{minipage}[t]{0.49\textwidth}
\vspace{0pt}
\centering
\small
\setlength{\tabcolsep}{5pt}
\renewcommand{\arraystretch}{0.85}

\caption{\footnotesize
\textbf{Performance under LIBERO-PRO perturbations.}
\emph{joint} denotes joint training of the video model
and action policy, whereas \emph{disj} denotes the
two-stage training procedure.}
\label{tab:perturbation_results}
\resizebox{\linewidth}{!}{%
\begin{tabular}{lccccc}
\toprule
\textbf{Approach}
& \textbf{None}
& \textbf{Obj.}
& \textbf{Pos.}
& \textbf{Sem.}
& \textbf{Task} \\
\midrule
IDM (joint)
& 97.9 & 83.7 & 2.7 & 96.3 & 23.7 \\

FastWAM (joint)
& 96.9 & 78.0 & 0.9 & 83.7 & 15.3 \\

\addlinespace[2pt]
\midrule
\addlinespace[2pt]

IDM (disj)
& 98.5 & 84.9 & 10.9 & 98.0 & 23.6 \\

FastWAM (disj)
& 97.6 & 79.5 & 10.5 & 94.7 & 20.4 \\

\addlinespace[2pt]
\midrule
\addlinespace[2pt]

\rowcolor{myblue}
1-Step from Prior
& 98.8 & \textbf{85.2} & \textbf{12.3}
& \textbf{98.5} & 23.5 \\

\rowcolor{myblue}
\textit{i}WAM
& \textbf{99.0} & 84.4 & 9.7
& 97.8 & \textbf{24.1 }\\
\bottomrule
\end{tabular}%
}
\end{minipage}

\end{table*}



\textbf{Baselines and Comparison.} For LIBERO-PRO, along with testing the original FastWAM implementation~\citep{yuan2026fastwam} that jointly trains the action policy and the video model,  we also adapt the two-stage training strategy of \citet{disjointvideo} to FastWAM and IDM, where we first fine-tune the video model independently, then freeze its parameters while training the action policy. In addition, we also consider the simplified 1-Step from Prior variant, which samples $z_1\sim p_{\mathrm{prior}}$ independently during training rather than using the noise endpoint associated with the demonstrated future. Full training and evaluation details are provided in Appendix~\ref{appendix:liberopro}.

\begin{wraptable}{R}{0.51\textwidth}
\vspace{-5pt}
\captionsetup{skip=3pt}
\vspace{-0.2cm}
\centering
\vspace{-0.3cm}
\scriptsize
\setlength{\tabcolsep}{2.6pt}
\renewcommand{\arraystretch}{0.95}
\caption{\footnotesize
\textbf{DROID simulation manipulation performance.}
Success rates (\%) over 20 trials per task at replanning
horizons of 16 and 24 action steps. The average is computed
across all reported task--horizon combinations.}
\label{tab:replanning_results}
\resizebox{\linewidth}{!}{%
\begin{tabular}{l cc cc cc c}
\toprule
& \multicolumn{2}{c}{Cube $\rightarrow$ Bowl}
& \multicolumn{2}{c}{Can $\rightarrow$ Mug}
& \multicolumn{2}{c}{Banana $\rightarrow$ Bin}
& \multirow{2}{*}{Avg.} \\
\cmidrule(lr){2-3}
\cmidrule(lr){4-5}
\cmidrule(lr){6-7}
Method
& 16 & 24
& 16 & 24
& 16 & 24
& \\
\midrule

FastWAM
& 0\% & 0\%
& 0\% & 0\%
& 20\% & 35\%
& 9.2\% \\

IDM
& \textbf{95\%} & \textbf{100\%}
& 95\% & 90\%
& \textbf{100\%} & \textbf{100\%}
& \textbf{96.7\%} \\

$\pi_{0.5}$\textsuperscript{$\dagger$}
& \textbf{95\%} & ---
& 55\% & ---
& 45\% & ---
& 65.0\% \\

IDM-flow map
& 90\% & 80\%
& 95\% & \textbf{95\%}
& \textbf{100\%} & 95\%
& 92.5\% \\

\midrule
\rowcolor{myblue}
\textit{i}WAM
& 80\% & 85\%
& \textbf{100\%} & \textbf{95\%}
& \textbf{100\%} & \textbf{100\%}
& 93.3\% \\

\rowcolor{myblue}
\textit{i}WAM (MIP)
& 90\% & 95\%
& 85\% & 85\%
& \textbf{100\%} & 95\%
& 91.7\% \\

\bottomrule
\end{tabular}%
}

\vspace{2pt}
\parbox{\linewidth}{%
\scriptsize
\textsuperscript{$\dagger$}$\pi_{0.5}$ uses its native action chunk of 15 actions and executes all 15 before replanning; its results are therefore listed under the 16-step column.
}
\vspace{-0.2cm}
\end{wraptable}
For DROID, we benchmark against two-stage IDM, which uses 8 flow-denoising steps at inference to generate the future. We also evaluate \emph{IDM-flow map}, which uses the flow map learned by \methodname{} to approximate the future in a single step. For FastWAM, we train both joint and two-stage variants, but omit the latter from the main results due to lower performance. All flow-matching action policies use 8 action-denoising steps, while \methodname{} (MIP) uses 2. For simulation experiments, we additionally evaluate the official DROID-finetuned $\pi_{0.5}$ checkpoint. Complete evaluation details are provided in Appendix~\ref{appendix:real_world_eval}.


\emph{\textbf{1) Can \methodname{} retain performance that roughly matches IDM in environments not seen during training, without explicitly generating future imaginations?}} On LIBERO-PRO, the prior-noise variant performs comparably to IDM and improves over FastWAM across perturbations (Table~\ref{tab:perturbation_results}). In unseen DROID simulation environments, \methodname{} achieves 93.3\% success compared with IDM's 96.7\%, while substantially outperforming FastWAM (9.2\%) and $\pi_{0.5}$ (65.0\%) (Table~\ref{tab:replanning_results}). On the physical robot, \methodname{} (MIP) achieves 85.0\% success, compared with 80.0\% for IDM and 13.3\% for FastWAM (Table~\ref{tab:real_world_results}).

\begin{figure}[t]
    \centering
    \vspace{-0.3cm}
    \begin{minipage}[t]{0.49\linewidth}
        \vspace{0pt}
        \includegraphics[width=\linewidth]{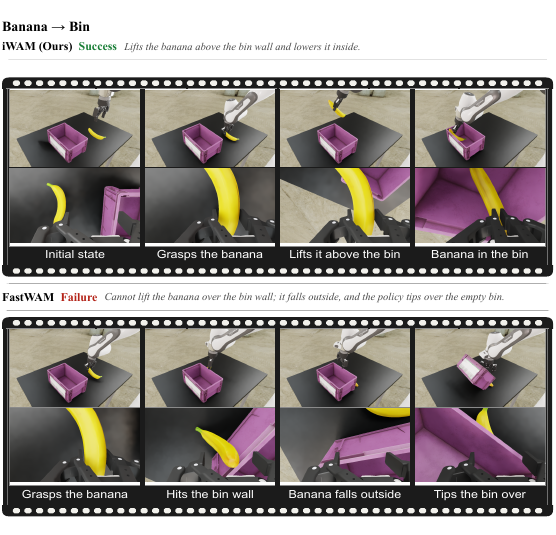}
    \end{minipage}\hfill
    \begin{minipage}[t]{0.49\linewidth}
        \vspace{0pt}
        \includegraphics[width=\linewidth]{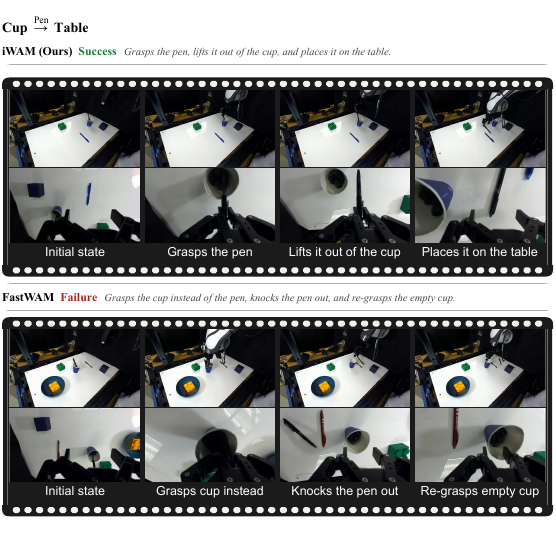}
    \end{minipage}
    \vspace{-0.4cm}
    \caption{\textbf{Evaluation behaviors in unseen environments.} FastWAM makes only partial progress on the banana-placement task (left) and exhibits poor instruction following on the pen-transfer task (right), whereas iWAM successfully completes both tasks. Additional rollouts are provided in Appendix~\ref{appendix:qualitative_rollouts}.}
    \label{fig:rollouts_main}
    \vspace{-0.3cm}
\end{figure}

Figure~\ref{fig:rollouts_main} illustrates the qualitative differences in unseen environments: \methodname{} completes the instructed tasks, whereas FastWAM only makes partial progress or manipulates the wrong object. Figure~\ref{fig:fastwam_failure_modes} shows that FastWAM can deviate from the instruction both at the start and after an unsuccessful attempt, switching to unrelated objects rather than retrying the intended manipulation. The strong performance of IDM in these unseen environments indicates that generating futures from the video model does provide effective conditioning for generalization, suggesting that FastWAM is unable to fully exploit the video model's generalization capabilities. \methodname{} retains these benefits by approximating computation needed for future conditioning in a single video-model evaluation, without explicitly generating the future.

\textbf{2) How does \methodname{} compare with IDM and FastWAM in task success and inference latency?}
We measure latency on a single NVIDIA A100 80 GB PCIe GPU at batch size 1, matching policy evaluation. We time all methods with two action-denoising steps, matching \methodname{} (MIP), so that all latency numbers 
\begin{wrapfigure}{r}{0.442\textwidth}
    \centering
    \includegraphics[width=\linewidth]{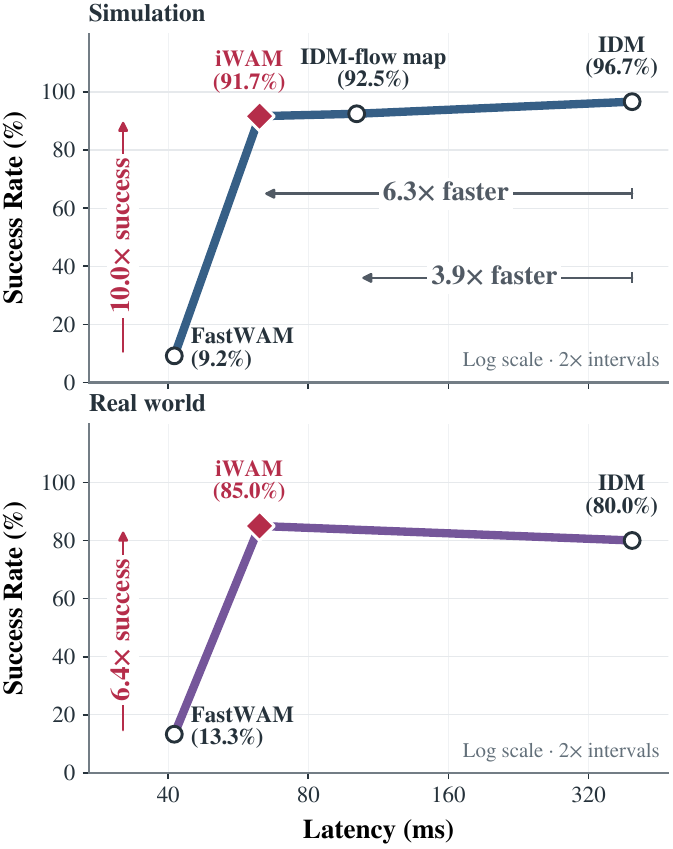}
    \captionsetup{font={footnotesize},labelfont={footnotesize}}
    \vspace{-0.6cm}
    \caption{{\textbf{Task success rates and inference latency on DROID.} Success is averaged over three tasks in simulation (top) and the real world (bottom), with simulation horizons as in Table~\ref{tab:replanning_results}. Latency uses two action-denoising steps for all methods; success uses each policy's evaluation budget (see the text to the left for details).}}
    \label{fig:perf_latency_full}
    \vspace{-0.7cm}
\end{wrapfigure}
reflect time needed for iterative video generation rather than more action-denoising steps. The success rates in Figure~\ref{fig:perf_latency_full} come from our main evaluations (Tables~\ref{tab:replanning_results} and~\ref{tab:real_world_results}), which use eight action-denoising steps for the flow-based WAMs and two for \methodname{} (MIP). 
We reduce flow-based action policies to two steps only for timing (i.e., IDM and FastWAM); their success rates are not recomputed at this budget and correspond to full 8-step action denoising. 

This timing protocol favors the baselines: FastWAM and IDM retain the success rates measured with eight action-denoising steps, while their latencies are measured using only two. In contrast, both the success rate and latency of \methodname{} (MIP) are measured with two steps. Thus, the protocol understates the baselines' latency relative to their evaluated configurations, making the reported speedup conservative. The baseline points should therefore be interpreted as somewhat \emph{optimistic} success-latency estimates, rather than measured two-step operating points.

Under this timing protocol, \methodname{} (MIP) has approximately $6\times$ lower latency than IDM; their simulation success rates are 91.7\% and 96.7\%, respectively. FastWAM is faster but achieves only 9.2\% success. IDM-flow map achieves 92.5\%, but requires two video-backbone evaluations: one to predict the future and a second to extract its conditioning activations. This additional evaluation makes it slower than \methodname{}, which obtains conditioning activations directly from noise in a single pass while achieving similar success. Thus, \methodname{} approaches IDM's task success with a single video-model pass, \mbox{at a modest latency cost over FastWAM.}

\textbf{3) When does training-time alignment with the demonstrated future matter most?} Our results support the requirements identified in \cref{sec:iwam}: exposing future-specific information with a single model evaluation, and aligning that information with demonstrated actions during training. On LIBERO-PRO, the 1-Step from Prior variant matches IDM and improves over FastWAM (Table~\ref{tab:perturbation_results}). This suggests that a single evaluation can provide useful predictive information even without explicitly matching the noise endpoint to the demonstration. The DROID results show the limits of this approach. Prior-noise conditioning yields substantially higher action-training loss and negligible task success (less than 3\% average success in DROID simulation), suggesting that access to predictive information alone is insufficient.

We attribute this difference to two factors. First, the more realistic DROID dataset exhibits substantially greater diversity in its demonstrated futures than LIBERO. The MSE between futures generated from randomly sampled noise and demonstrated futures is over $10\times$ higher on DROID than on LIBERO (Appendix~\ref{appendix:prior_noise_mse}). Thus, LIBERO's more constrained task and demonstration distribution make randomly sampled futures more consistent with demonstrations, leaving limited additional benefit from inversion. Second, DROID presents a substantially larger evaluation shift. LIBERO-PRO applies controlled perturbations to training scenes, whereas our DROID evaluations take place in environments entirely unseen during training. As shown in \cref{sec:liberopro}, FastWAM remains relatively close to IDM under the former but falls far behind under the latter. This suggests that the limitations of policies that do not effectively leverage future conditioning become more pronounced under larger distribution shifts. Together, these observations suggest that alignment and future-specific conditioning are particularly important when demonstrations are diverse and evaluation conditions differ substantially from training.

In these settings, the bidirectional flow map provides the required alignment during training by approximately recovering the noise endpoint associated with the demonstrated future. A single noise-to-clean evaluation at this endpoint supplies activations that represent the same future as the action supervision, which makes the correspondence between conditioning and actions easier to learn. The reduction in training loss toward IDM's loss (Figure~\ref{fig:ifc_training_loss}), together with the recovery in task success, supports this explanation. \methodname{}'s competitive performance further suggests that these activations retain the future-specific information needed for action prediction without requiring an accurate reconstruction of the full future.

%% file: sections/conclusion.tex
\vspace{-0.2cm}
\section{Discussion and Future Work}
\label{sec:conclusion}
\vspace{-0.2cm}

We presented \methodname{}, which conditions an action policy on the intermediate features of a single video-model forward pass rather than an explicitly generated future. Our approach builds on a simple observation: uniquely identifying an action from a video model's features of the current observation requires explicit test-time compute. We show that this is one of the reasons why methods that do not spend any test-time compute (e.g., FastWAM) fail to generalize. However, by design, standard flow-matching models underutilize test-time compute when restricted to a single video-model evaluation. To address this, we trained a bidirectional flow map and used a single forward pass through it to predict actions. Across simulated and real-world manipulation tasks, \methodname{} matches the performance of world action models that explicitly generate the future at a fraction of their inference cost, while substantially outperforming methods that condition only on the current observation. These results suggest that policies can benefit from the ability to imagine the full future without paying the cost of generating it.

The ideas behind \methodname{} could also make model-based planning more efficient. Planners built on video models typically generate candidate futures explicitly and score each one with a value function~\citep{cosmospolicy}. Analogous to our action policy, the value function could instead be conditioned on intermediate activations from a single noise-to-clean evaluation of the bidirectional flow map at a sampled noise endpoint, without generating the corresponding future. As in policy training, alignment would remain necessary: a clean-to-noise evaluation would recover the noise endpoint associated with each demonstrated future, so that the conditioning represents the same future as the value target used for supervision. Finally, rather than training a separate value network, the bidirectional flow map itself could be trained to predict the value jointly with the average velocity. This would enable efficient best-of-$N$ planning: sampling $N$ noise endpoints, scoring each with a single model evaluation in parallel, and selecting the highest-value candidate without explicitly generating any futures.

Another promising direction is to strengthen the bidirectional flow map with more advanced distillation objectives. Despite using a relatively simple shortcut-style objective, \methodname{} already nearly matches IDM, leaving only a very small performance gap. We adopted shortcut-style training because it is simple, stable when predicting many frames jointly, and avoids computing network derivatives. The small remaining gap likely reflects approximation error in the learned flow map, suggesting that stronger distillation objectives could close it entirely. Recent distillation methods such as rCM~\citep{rcm} and its causal variant~\citep{crcm} achieve high-fidelity few-step video generation at scale and are natural candidates. These objectives could be complemented with an auxiliary adversarial loss, as in One-Forcing~\citep{oneforcing}, potentially using the bidirectionally attentive video discriminator of AAD-1~\citep{aad1}. Any such objective would need to be extended to both transport directions, since \methodname{} also relies on the clean-to-noise direction to recover noise endpoints aligned with demonstrated futures.

\vspace{-0.2cm}
\section*{Acknowledgements}
\vspace{-0.2cm}
We thank Yutong He, Sandeep Routray, Gene Yang, and members of the CMU AIRe Lab for discussions and feedback. We also thank Oliver Kroemer for generously providing access to the Franka Panda robotic arm and experimental apparatus used in our DROID experiments. This work is supported by Office of the Naval Research (ONR) through N0014-24-12206 and partly through the Toyota Research Institute U3.0 program. We thank the Google TPU Research Cloud (TRC) program for providing the computational resources that supported this work.


%% file: sections/appendix.tex
\section{Preliminaries: Shortcut Models \& Guidance}
\label{appendix:shortcut}
 \textbf{Classifier-free guidance and guidance-free training.}
Conditional generation commonly uses classifier-free guidance (CFG) to strengthen adherence to the conditioning input $ y$, providing explicit prompt
guidance that can be important for test-time generalization. Let $v^\tau_{\theta_v}(z_\tau\mid\varnothing)$ denote the unconditional marginal velocity, where $\varnothing$ denotes dropped conditioning. At guidance strength $w\geq1$, CFG forms the guided velocity
\begin{align}
v^\tau_{\theta_v}(z_\tau\mid y,w)
&=
w\,v^\tau_{\theta_v}(z_\tau\mid y)
-
(w-1)\,v^\tau_{\theta_v}(z_\tau\mid\varnothing).
\label{eq:cfg}
\end{align}
Therefore, it requires separate conditional and unconditional model evaluations at each sampling step. 
Guidance-Free Training (GFT)~\citep{gft} instead trains a single network to directly predict the prompt-guided marginal velocity. Rearranging Eq.~\eqref{eq:cfg} gives
\begin{align}
v^\tau_{\theta_v}(z_\tau\mid y)
&=
\frac{1}{w}\,v^\tau_{\theta_v}(z_\tau\mid y,w)
+
\left(1-\frac{1}{w}\right)
v^\tau_{\theta_v}(z_\tau\mid\varnothing),
\label{eq:gft_reparam}
\end{align}
which allows the guided field to be trained using the original flow-matching target:
\begin{align}
\mathcal{L}_{\mathrm{GFT}}
&=
\mathbb{E}\!\left[
\left\|
\frac{1}{w}\,v^\tau_{\theta_v}(z_\tau\mid y,w)
+
\left(1-\frac{1}{w}\right)
\operatorname{sg}\!\left[
v^\tau_{\theta_v}(z_\tau\mid\varnothing)
\right]
-
u_\tau
\right\|_2^2
\right],
\label{eq:gft}
\end{align}
where $\operatorname{sg}[\cdot]$ denotes stop-gradient. Sampling $w$ during training and providing it as an input allows a single network to predict guided velocities across guidance strengths. Importantly for our setting, GFT also produces internal activations corresponding directly to the guided velocity in a single forward pass, which we exploit in Section~\ref{sec:iwam_practice}.

\textbf{Shortcut models.}
Shortcut Models~\citep{shortcut} learn to transport samples between distant flow times by conditioning on flow time and step size, thereby supporting variable sampling budgets. They can be viewed as a progressive self-distillation scheme, where flow matching provides supervision for infinitesimal steps and larger steps are bootstrapped from smaller ones~\citep{flowmaps}. Existing shortcut model formulations, however, focus on noise-to-clean transport and do not provide the reverse clean-to-noise mapping needed to associate a clean sample with its noise endpoint. Moreover, Shortcut Models report that standard CFG becomes unreliable for large steps and requires the guidance scale to be specified during training. In our work, we will extend this formulation to arbitrary two-time transport and directly learned guided velocities, enabling bidirectional transport and arbitrary guidance strengths in the single-step regime.

\begin{proposition}[Marginal velocity at the noise endpoint]
\label{proof:noise_endpoint_velocity}
Let $c=(o_0,\ell)$ denote the conditioning context and consider the linear
interpolant $z_\tau=(1-\tau)o_{1:N}+\tau\epsilon,$ where $\epsilon \sim p_{\mathrm{prior}}$ is sampled independently of $(c,o_{1:N})$. Then, the
marginal vector field at the noise endpoint $\tau=1$ satisfies
\begin{align}
    v^1(z_1\mid o_0,\ell)
    &=
    z_1-\mathbb{E}\!\left[o_{1:N}\mid o_0,\ell\right].
    \label{eq:noise_endpoint_velocity}
\end{align}
Consequently, a one-step denoising estimate obtained from the endpoint
velocity is
\begin{align}
    z_1-v^1(z_1\mid c)
    &=
    \mathbb{E}\!\left[o_{1:N}\mid c\right],
    \label{eq:one_step_future}
\end{align}
which is independent of the sampled noise $z_1$.
\end{proposition}

\begin{proof}
For the linear interpolant, the conditional velocity is
\begin{align}
    u_\tau &= \frac{d z_\tau}{d\tau}=\epsilon-o_{1:N}.
\end{align}
Hence, the marginal vector field is
\begin{align}
    v^\tau(z_\tau\mid c)
    &=
    \mathbb{E}\!\left[
        \epsilon-o_{1:N}\mid z_\tau,c
    \right].
\end{align}
At $\tau=1$, $z_1=\epsilon$. Since $\epsilon$ is independent of
$(c,o_{1:N})$,
\begin{align}
    v^1(z_1\mid c)
    &=
    \mathbb{E}\!\left[
        z_1-o_{1:N}\mid z_1,c
    \right] \\
    &=
    z_1-\mathbb{E}\!\left[o_{1:N}\mid c\right].
\end{align}
Rearranging gives Eq.~\eqref{eq:one_step_future}.
\end{proof}

\section{Detailed Analysis on LIBERO-PRO}
\label{appendix:liberopro}

\subsection{Evaluation on LIBERO-PRO.}
 As in Section~\ref{sec:liberopro}, we train all models only on the original LIBERO demonstrations and evaluate them zero-shot, without further fine-tuning, on both the unperturbed tasks and the four LIBERO-PRO perturbation types defined in Table~\ref{tab:perturbation_results_full}. For each suite and evaluation condition, we run 50 trials for each of the 10 tasks using the benchmark's fixed initial states, yielding 500 trials per table entry. The Average columns aggregate results across the four suites, corresponding to 2{,}000 trials per evaluation condition. Following~\citet{yuan2026fastwam}, we use a batch size of 128 and train all models for 10 epochs. For two-stage variants, both the video model and the action policy are trained for 10 epochs each. At evaluation, we likewise adopt FastWAM's inference configuration and apply it identically across all methods.

 The original FastWAM implementation~\citep{yuan2026fastwam} jointly trains the action policy and the video model, allowing gradients from the action loss to update the pretrained video model. Due to the narrow action-training data distribution of LIBERO, we find that this can substantially degrade the generalization performance inherited from video pretraining. We adapt the two-stage training strategy of \citet{disjointvideo} to FastWAM and IDM, where we first fine-tune the video model independently, then freeze its parameters while training the action policy. This substantially improves FastWAM and IDM’s performance across perturbations. However, even with this training strategy, across perturbation categories (see Table~\ref{tab:perturbation_results_full}), we find that FastWAM exhibits significantly worse generalization performance than IDM. In addition, we consider the simplified 1-Step from Prior variant, which samples $z_1\sim p_{\mathrm{prior}}$ independently during training rather than using the noise endpoint associated with the demonstrated future, while using the same Stage~1 video backbone.

\subsection{Performance on LIBERO-PRO.}
As shown in Table~\ref{tab:perturbation_results_full}, FastWAM generalizes worse than IDM across all perturbation categories. In particular, there is a significant gap under object-size perturbations on LIBERO-Goal, where IDM achieves 60.8\% compared to 44.4\% for FastWAM (Table~\ref{tab:libero_goal_object_aggregate}); detailed task-level results are provided in Table~\ref{tab:libero_goal_object_task_results}. When the grasped object is enlarged, IDM’s imagined future correctly captures the gripper interacting with the larger object, while FastWAM’s action chunk does not appear to appropriately account for the increase in object size. This suggests that future-predictive information provides geometric cues important for generalization under distribution shift.

\begin{table*}[h]
    \centering
    \caption{\textbf{Performance under LIBERO-PRO perturbations.} Models are trained on LIBERO and evaluated under perturbations to object attributes (Obj.), initial object positions (Pos.), instruction phrasing while preserving task intent (Sem.), and instructions specifying a different target object or action (Task); None denotes the unperturbed setting. \emph{joint} denotes joint training of the video model and action policy, whereas \emph{disj} denotes the two-stage training procedure.}
    \label{tab:perturbation_results_full}

    \setlength{\tabcolsep}{3.8pt}
    \renewcommand{\arraystretch}{1.15}
    \small

    \begin{adjustbox}{max width=\textwidth}
    \begin{tabular}{l*{25}{c}}
        \toprule

        \multirow{2}{*}{\textbf{Approach}}
        & \multicolumn{5}{c}{\textbf{LIBERO-10}}
        & \multicolumn{5}{c}{\textbf{LIBERO-Goal}}
        & \multicolumn{5}{c}{\textbf{LIBERO-Object}}
        & \multicolumn{5}{c}{\textbf{LIBERO-Spatial}}
        & \multicolumn{5}{c}{\textbf{Average}} \\

        \cmidrule(lr){2-6}
        \cmidrule(lr){7-11}
        \cmidrule(lr){12-16}
        \cmidrule(lr){17-21}
        \cmidrule(l){22-26}

        & \textbf{None} & \textbf{Obj.} & \textbf{Pos.} & \textbf{Sem.} & \textbf{Task}
        & \textbf{None} & \textbf{Obj.} & \textbf{Pos.} & \textbf{Sem.} & \textbf{Task}
        & \textbf{None} & \textbf{Obj.} & \textbf{Pos.} & \textbf{Sem.} & \textbf{Task}
        & \textbf{None} & \textbf{Obj.} & \textbf{Pos.} & \textbf{Sem.} & \textbf{Task}
        & \textbf{None} & \textbf{Obj.} & \textbf{Pos.} & \textbf{Sem.} & \textbf{Task} \\

        \midrule

        \textbf{IDM (joint)}
        & 96.4 & 66.6 & 0.4 & 95.4 & 16.2
        & 97.4 & 79.0 & 8.2 & 97.4 & 9.0
        & 98.6 & 92.4 & 0.0 & 95.0 & \textbf{10.2}
        & \textbf{99.4} & 96.6 & 2.2 & 97.4 & \textbf{59.4}
        & 97.9 & 83.7 & 2.7 & 96.3 & 23.7 \\

        \textbf{FastWAM (joint)}
        & 94.6 & 51.6 & 0.0 & 80.8 & 10.6
        & 96.6 & 71.2 & 3.0 & 93.4 & 8.6
        & 99.4 & 93.4 & 0.0 & 99.2 & 9.4
        & 97.0 & 95.8 & 0.6 & 61.4 & 32.6
        & 96.9 & 78.0 & 0.9 & 83.7 & 15.3 \\

        \textbf{IDM ($p=0$)}
        & 97.2 & 64.8 & 6.6 & 96.4 & \textbf{20.8}
        & 98.4 & 79.0 & 11.4 & 98.0 & \textbf{12.8}
        & 99.2 & \textbf{96.2} & \textbf{9.8} & 97.8 & 10.0
        & 98.4 & 98.2 & 10.8 & 96.6 & 53.0
        & 98.3 & 84.5 & 9.7 & 97.2 & \textbf{24.1} \\
        \addlinespace[2pt]
        \midrule
        \addlinespace[2pt]

        \textbf{IDM (disj)}
        & 97.2 & \textbf{68.6} & 4.2 & 97.2 & 20.0
        & 98.0 & 79.8 & 17.2 & \textbf{98.4} & 10.6
        & \textbf{99.6} & 94.0 & 8.2 & 98.6 & 10.0
        & \textbf{99.4} & 97.0 & 14.2 & 97.6 & 53.8
        & 98.5 & 84.9 & 10.9 & 98.0 & 23.6 \\

        \textbf{FastWAM (disj)}
        & 96.0 & 57.0 & 4.8 & 92.2 & 15.6
        & 97.4 & 73.4 & 11.8 & 95.8 & 10.4
        & 99.4 & 91.0 & 3.2 & \textbf{99.8} & 10.0
        & 97.8 & 96.6 & \textbf{22.2} & 91.0 & 45.4
        & 97.6 & 79.5 & 10.5 & 94.7 & 20.4 \\

        \addlinespace[2pt]
        \midrule
        \addlinespace[2pt]
        \rowcolor{myblue}
        \textbf{1-Step from Prior}
        & \textbf{98.0} & 64.4 & \textbf{8.2} & \textbf{98.6} & 19.2
        & \textbf{98.6} & \textbf{83.0} & \textbf{17.4} & 98.0 & 11.0
        & 99.4 & 94.8 & 8.0 & 99.6 & 10.0
        & \textbf{99.4} & \textbf{98.4} & 15.6 & \textbf{97.8} & 53.8
        & \textbf{98.8} & \textbf{85.2} & \textbf{12.3} & \textbf{98.5} & 23.5 \\

        \bottomrule
    \end{tabular}
    \end{adjustbox}
\end{table*}

Furthermore, FastWAM and several IDM variants jointly train the action expert and video model, allowing the action loss to backpropagate through the video model. Fine-tuning a pretrained video model in this manner may degrade its generalization, particularly when the action-training data are narrowly distributed. To investigate this hypothesis, we decouple the two training stages: we first fine-tune the video model independently, freeze its parameters, and then train the action expert using the action loss. We apply this disjoint training procedure to both FastWAM and IDM, denoted FastWAM (disj) and IDM (disj), respectively, with results reported in Table~\ref{tab:perturbation_results_full}.

Disjoint training yields substantial gains under initial-state perturbations, improving IDM from 2.7\% to 10.9\% and FastWAM from 0.9\% to 10.5\%, while also improving performance under the other perturbation categories. Qualitative analysis of the predicted videos indicates that these gains stem from more accurate future predictions, particularly for initial-state perturbations, where the jointly trained IDM video model tends to overfit to the training-time initial object positions. FastWAM also benefits substantially under object-size and language perturbations, improving from 83.7\% to 94.7\% under language perturbations and from 44.4\% to 53.6\% on LIBERO-Goal object-resize perturbations (Table~\ref{tab:libero_goal_object_aggregate}). 

To better understand why joint training hurts IDM, we examine the role of future-observation augmentation during action training. In the standard joint setting, the future observations provided to the action expert are randomly noised to a sampled noise level with probability $0.5$. When we remove this augmentation ($p=0$) and always provide clean futures, joint IDM recovers performance comparable to disjoint training across perturbations (Table~\ref{tab:perturbation_results_full}). This suggests that, when clean futures are available, the action expert can fit the action objective with substantially less modification to the shared video-model representations, thereby better preserving the generalization inherited from pretraining despite action-loss backpropagation. 

This also helps explain why FastWAM continues to benefit from disjoint training on LIBERO. Without future-derived conditioning, its action expert must infer the action sequence from the current observation and instruction alone. Optimizing the action objective therefore places greater pressure on the shared video-model representations, making them more susceptible to specialization to LIBERO's relatively narrow action-training distribution. Freezing the video model prevents this adaptation and preserves more of its pretrained generalization, leading to the stronger robustness observed for FastWAM (disj).


\begin{table}[t]
\centering
\caption{Aggregate success rates under LIBERO-Goal object perturbations.
Results are averaged over 5 tasks per perturbation type, with 50 trials per task.
The resize setting combines a $1.5\times$ bowl size change with color/appearance changes.}
\label{tab:libero_goal_object_aggregate}

\setlength{\tabcolsep}{7.0pt}
\renewcommand{\arraystretch}{1.15}
\small
\begin{tabular}{lcccc}
    \toprule
    \textbf{Perturbation Type}
    & \textbf{FastWAM-joint}
    & \textbf{IDM-joint}
    & \textbf{FastWAM-disj}
    & \textbf{IDM-disj} \\
    \midrule
    Color / appearance only
    & \textbf{98.0}
    & 97.2
    & 93.2
    & 94.4 \\
    Object resize ($1.5\times$ bowl) + color / appearance
    & 44.4
    & 60.8
    & 53.6
    & \textbf{65.2} \\
    \midrule
    Average
    & 71.2
    & 79.0
    & 73.4
    & \textbf{79.8} \\
    \bottomrule
\end{tabular}

\end{table}

\begin{table*}[t]
\centering
\caption{Success rates under object perturbations on LIBERO-Goal.
Object substitutions modify either appearance only or both appearance and object size.
Higher values indicate better performance.}
\label{tab:libero_goal_object_task_results}

\setlength{\tabcolsep}{5.0pt}
\renewcommand{\arraystretch}{1.12}
\small
\begin{adjustbox}{max width=\textwidth}
\begin{tabular}{clllcccc}
    \toprule
    \textbf{\#}
    & \textbf{Task}
    & \textbf{Object Substitution}
    & \textbf{Size Change}
    & \textbf{FastWAM-joint}
    & \textbf{IDM-joint}
    & \textbf{FastWAM-disj}
    & \textbf{IDM-disj} \\
    \midrule
    0
    & \texttt{open\_the\_middle\_drawer\_of\_the\_cabinet}
    & wooden cabinet $\rightarrow$ yellow cabinet
    & No
    & \textbf{100.0}
    & 96.0
    & \textbf{100.0}
    & 86.0 \\
    1
    & \texttt{put\_the\_bowl\_on\_the\_stove}
    & bowl $\rightarrow$ bigger bowl; stove $\rightarrow$ yellow stove
    & Bowl $1.5\times$
    & 64.0
    & 92.0
    & 82.0
    & \textbf{94.0} \\
    2
    & \texttt{put\_the\_wine\_bottle\_on\_top\_of\_the\_cabinet}
    & wine bottle $\rightarrow$ white bottle; cabinet $\rightarrow$ yellow cabinet
    & No
    & \textbf{96.0}
    & 92.0
    & 74.0
    & 86.0 \\
    3
    & \texttt{open\_the\_top\_drawer\_and\_put\_the\_bowl\_inside}
    & cabinet $\rightarrow$ yellow cabinet; bowl $\rightarrow$ bigger bowl
    & Bowl $1.5\times$
    & 10.0
    & 18.0
    & 14.0
    & \textbf{32.0} \\
    4
    & \texttt{put\_the\_bowl\_on\_top\_of\_the\_cabinet}
    & bowl $\rightarrow$ bigger bowl; cabinet $\rightarrow$ yellow cabinet
    & Bowl $1.5\times$
    & 32.0
    & 46.0
    & 44.0
    & \textbf{48.0} \\
    5
    & \texttt{push\_the\_plate\_to\_the\_front\_of\_the\_stove}
    & plate $\rightarrow$ yellow plate
    & No
    & 98.0
    & \textbf{100.0}
    & \textbf{100.0}
    & \textbf{100.0} \\
    6
    & \texttt{put\_the\_cream\_cheese\_in\_the\_bowl}
    & bowl $\rightarrow$ bigger bowl; cream cheese $\rightarrow$ red cream cheese
    & Bowl $1.5\times$
    & 56.0
    & \textbf{78.0}
    & 68.0
    & 74.0 \\
    7
    & \texttt{turn\_on\_the\_stove}
    & stove $\rightarrow$ yellow stove
    & No
    & \textbf{100.0}
    & \textbf{100.0}
    & 98.0
    & \textbf{100.0} \\
    8
    & \texttt{put\_the\_bowl\_on\_the\_plate}
    & bowl $\rightarrow$ bigger bowl; plate $\rightarrow$ yellow plate
    & Bowl $1.5\times$
    & 60.0
    & 70.0
    & 60.0
    & \textbf{78.0} \\
    9
    & \texttt{put\_the\_wine\_bottle\_on\_the\_rack}
    & wine bottle $\rightarrow$ white bottle; wine rack $\rightarrow$ brown rack
    & No
    & 96.0
    & 98.0
    & 94.0
    & \textbf{100.0} \\
    \bottomrule
\end{tabular}
\end{adjustbox}

\end{table*}

\section{Detailed Experimental setup of DROID}
\subsection{Training Details}
\label{appendix:droid_training}
\paragraph{Data and preprocessing.}
We train on DROID~\citep{droid} using the DreamZero release~\citep{dreamzero}, which removes idle frames and unlabeled or unsuccessful episodes, yielding 57,774 episodes and 14.7M frames at 15\,Hz. Each training example consists of the current observation, a 32-step action chunk ($\approx$2.1\,s), and the corresponding future video. We subsample the future by $2\times$ in time to obtain 16 future frames; together with the current frame, this gives 17 frames, which the Wan2.2 VAE compresses into 5 latent frames. We use the wrist and first exterior camera, resize each view to $160\times320$, and concatenate them side by side before VAE encoding. Actions contain seven joint-position targets, expressed as deltas relative to the current joint state, and one absolute gripper command. We normalize actions to $[-1,1]$ using the 1st and 99th percentiles, with separate statistics for each action dimension and each of the 32 positions within the chunk. Proprioception consists of the current seven joint positions and gripper state and is provided only to the action expert; the video model receives no proprioceptive input. We apply temporally consistent color jitter following DreamZero~\citep{dreamzero}, sample one language annotation per episode when multiple are available, and precompute instruction embeddings using the frozen UMT5 encoder from Wan2.2.

\paragraph{Training procedure.}
We initialize the video model from Wan2.2-TI2V-5B, keeping the VAE and text encoder frozen. Training proceeds in two stages. In Stage~1, we fine-tune the full video DiT for 100K steps with a global batch size of 256 ($\sim$1.7 epochs) using the bidirectional shortcut objective described in Section~\ref{sec:iwam}, together with Guidance-Free Training~\citep{gft}. We use a maximum number of sampling steps $K_{\max}=32$ for the shortcut training objective, sample the inverse-guidance parameter $\beta\sim\mathcal{U}(0,1)$, and apply 10\% conditioning dropout. Three quarters of each batch use the standard flow-matching loss, while the remaining quarter uses the Shortcut bootstrap loss, with noise-to-clean and clean-to-noise transport sampled equally. Bootstrap targets are constructed from two half-steps using a Polyak-EMA target network with update rate $10^{-3}$. We additionally apply a round-trip consistency loss to one eighth of each batch, with its weight warmed up to 0.2 following the bidirectional training principle of Bidirectional Consistency Models~\citep{bcm}. Flow times are sampled using the shifted Wan2.2 schedule with shift factor 5, which biases training toward higher-noise flow times. When initialized without GFT, shortcut-style training incurs only about 16\% additional cost over an equivalent flow model~\citep{shortcut}.

In Stage~2, we initialize the frozen video model from the Stage~1 EMA weights and train a freshly initialized action policy for 100K steps with a batch size of 256. Unlike Stage~1, we do not use a shifted flow-time schedule for action training; instead, flow times are sampled uniformly from $[0,1]$. For \methodname{}, we recover the noise endpoint of each demonstrated future using two clean-to-noise flow-map steps at guidance strength $w=7$. The subsequent noise-to-clean evaluation that conditions the action policy uses the same guidance strength, both during training and at inference. The action policy uses 30 transformer layers, hidden dimension $d_a=1024$, and FFN dimension 4096, totaling approximately 1B trainable parameters. Since the learned flow map also serves as a standard flow model, we reuse the same trained video model for both \methodname{} and IDM to ensure a controlled comparison. For both methods, we freeze the video model and train their respective action policies for 100K steps; during IDM evaluation, the shared video model is used with standard flow integration. \emph{IDM-flow map}  also uses the learned flow map to approximate the future in a single step during evaluation.

For FastWAM, we follow the same official architecture and training procedure described above, using the same data, preprocessing, action normalization, model architecture, and optimizer as \methodname{}. Specifically, we jointly train a separate video model and action policy for 200K steps with a global batch size of 256, as increasing training from 100K to 200K steps slightly improves performance before saturating. FastWAM trains the video model using the standard flow-matching loss rather than the bidirectional shortcut objective used for \methodname{}.

The FastWAM (disj) variant was omitted from the main results because it underperformed the official jointly trained version; its task-level scores on DROID simulation are reported in Table~\ref{tab:fastwam_disj_results}. For the two-stage training procedure of FastWAM, we use the same video checkpoint as iWAM and follow the same Stage~2 training procedure outlined above, except that the action policy conditions only on the activations of the current observation.

\begin{wraptable}[11]{r}{0.51\textwidth}
\vspace{-0.5cm}
\centering
\scriptsize
\setlength{\tabcolsep}{2.6pt}
\renewcommand{\arraystretch}{0.95}
\captionsetup{font={footnotesize}}
\caption{
\textbf{DROID Simulation manipulation performance.}
Comparison of FastWAM, its two-stage variant FastWAM (disj), and iWAM. Success rates (\%) are measured over 20 trials per task at replanning horizons of 16 and 24 action steps. The average is computed across all task--horizon combinations.
}
\label{tab:fastwam_disj_results}
\vspace{-0.2cm}

\resizebox{\linewidth}{!}{%
\begin{tabular}{l cc cc cc c}
\toprule
& \multicolumn{2}{c}{Cube $\rightarrow$ Bowl}
& \multicolumn{2}{c}{Can $\rightarrow$ Mug}
& \multicolumn{2}{c}{Banana $\rightarrow$ Bin}
& \multirow{2}{*}{Avg.} \\
\cmidrule(lr){2-3}
\cmidrule(lr){4-5}
\cmidrule(lr){6-7}
Method
& 16 & 24
& 16 & 24
& 16 & 24
& \\
\midrule

FastWAM
& 0\% & 0\%
& 0\% & 0\%
& 20\% & 35\%
& 9.2\% \\

FastWAM (disj)
& 0\% & 0\%
& 0\% & 0\%
& 0\% & 10\%
& 1.7\% \\

\midrule
\rowcolor{myblue}
iWAM
& \textbf{80\%} & \textbf{85\%}
& \textbf{100\%} & \textbf{95\%}
& \textbf{100\%} & \textbf{100\%}
& \textbf{93.3\%} \\

\bottomrule
\end{tabular}%
}
\vspace{-0.4cm}
\end{wraptable}

All flow-based action policies are trained with the flow-matching formulation. We additionally train an \methodname{} variant using the Minimal Iterative Policy (MIP) objective~\citep{mip}, which predicts actions in two denoising steps: an initial regression step followed by refinement; we denote this variant as \methodname{} (MIP). We also evaluated a two-stage FastWAM variant, but it underperformed the jointly trained version and is therefore omitted from the main results.

\paragraph{Optimization.}
Both stages use AdamW with $(\beta_1,\beta_2)=(0.9,0.95)$, $\epsilon=10^{-8}$, weight decay $10^{-5}$, gradient clipping at norm 1.0, and a global batch size of 256. For Stage~1 bidirectional Shortcut training, the learning rate is linearly warmed up for 5K steps to $10^{-5}$ and then kept constant. For Stage~2, the learning rate warms up for 5K steps to $10^{-4}$ and then follows cosine decay to $10^{-6}$. Unless otherwise noted, Stage~2 models are trained for 100K steps; the FastWAM baseline is trained for 200K steps with the cosine schedule extended accordingly. Training uses bfloat16 computation with fp32 master weights and optimizer states.

\subsection{Consistency of Randomly Sampled Futures with Demonstrations}
\label{appendix:prior_noise_mse}

We measure how closely futures generated from randomly sampled noise match the demonstrated futures on LIBERO and DROID to assess differences in future diversity across the two datasets.

\textbf{Setup.} For each training sample, we generate a future from randomly sampled Gaussian noise using 1, 2, and 4 denoising steps using a standard flow model. We report the MSE between the decoded reconstruction and the VAE-decoded ground truth, averaged over future-frame RGB pixels and all samples, excluding the current observation. Both videos are in the $[0,1]$ RGB range when computing the MSE.

\begin{table}[h]
    \centering
    \caption{\textbf{MSE between futures generated from randomly sampled noise and demonstrated futures.} Values are reported in units of $10^{-3}$.}
    \label{tab:prior_noise_mse}
    \begin{tabular}{lccc}
        \toprule
        Dataset & 1-step & 2-step & 4-step \\
        \midrule
        LIBERO & 2.225 & 2.213 & 2.111 \\
        DROID  & 27.710 & 30.801 & 32.898 \\
        \bottomrule
    \end{tabular}
\end{table}

\textbf{Results.} Across all denoising step counts, the MSE on DROID is more than ten times higher than on LIBERO (Table~\ref{tab:prior_noise_mse}). Pixel-level MSE is only a coarse proxy, since it also reflects differences in visual content between the two datasets. Nevertheless, the size of the gap is consistent with LIBERO's more constrained task and demonstration distribution making randomly sampled futures more consistent with the demonstrations, which leaves limited additional benefit from inversion.

\subsection{Joint vs. Disjoint Training Across LIBERO and DROID}
\label{appendix:joint_disjoint_analysis}
As shown in Table~\ref{tab:fastwam_disj_results}, FastWAM (disj) achieves an average success rate of 1.7\%, compared to 9.2\% for jointly trained FastWAM. This contrasts with LIBERO, where disjoint training substantially improves FastWAM's robustness under perturbations (Table~\ref{tab:perturbation_results_full}). We hypothesize that this reversal reflects DROID's substantially greater diversity of demonstrated futures compared to LIBERO. Since FastWAM does not condition on a specific future, in particular the one realized by the supervised actions, learning to predict actions across DROID's more diverse demonstrations may benefit more from adapting the shared video-model representations to the action objective. Joint training enables this adaptation, whereas freezing the video model restricts it and can constrain action learning. Thus, while freezing helps preserve the pretrained generalization of the video model on LIBERO's narrower action-training distribution, the benefits of action-relevant adaptation may outweigh those of preserving these pretrained representations for FastWAM on DROID.

Together with the LIBERO analysis, these findings suggest that the benefits of preserving pretrained video-model representations depend on the information available to the action expert. For IDM-style approaches that condition action prediction on a specific future, preserving pretrained predictive capabilities is particularly important. In contrast, approaches without such conditioning, such as FastWAM, may benefit from adapting the shared representations to the action objective, particularly on more diverse demonstration distributions. More broadly, these results underscore the importance of exposing the action expert to future-predictive information at test time for robust generalization.

\subsection{Benchmark Initialization}
\label{appendix:init}
For DROID, we benchmark against IDM trained using the two-stage procedure, which uses 8 flow-denoising steps at inference to generate the future, requiring 9 video-backbone forward passes in total. We also evaluate a variant that uses the flow map learned by \methodname{} to approximate the future in a single step, which we denote as \emph{IDM-flow map}, reducing action prediction to 2 video-backbone forward passes. For FastWAM, we train both joint and two-stage variants, but omit the latter from the main results due to its relatively low performance. For both simulation and real-world evaluations, we use a guidance strength of $w=7$ for \methodname{} and both IDM variants; FastWAM does not have a corresponding guidance-strength parameter at inference. All flow-matching action policies use 8 action-denoising steps, and iWAM (MIP) uses 2 action-denoising steps. For simulation experiments, we additionally evaluate the official DROID-finetuned $\pi_{0.5}$ checkpoint using its default 10-step action-denoising procedure.


\begin{figure}[!t]
    \centering
    \begin{minipage}[t]{0.43\linewidth}
        \vspace{0pt}
        \includegraphics[width=\linewidth]{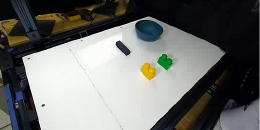}
    \end{minipage}\hfill
    \begin{minipage}[t]{0.43\linewidth}
        \vspace{0pt}
        \includegraphics[width=\linewidth]{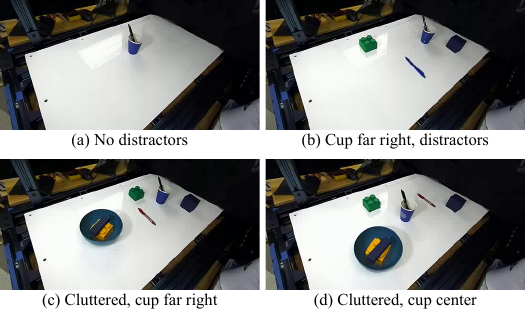}
    \end{minipage}\par\smallskip
    \includegraphics[width=0.43\linewidth]{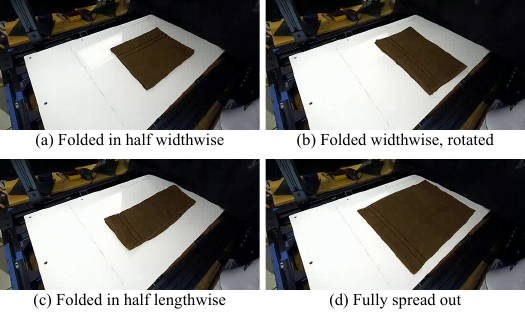}
    \caption{\textbf{Initial configurations for real-world evaluation.} The block-placement task starts with three blocks and a bowl on the table (top left). The pen-transfer task varies the cup placement and surrounding distractors (top right), while the towel-folding task varies the towel's fold state and orientation (bottom).}
    \label{fig:real_world_initial_states}
\end{figure}

\subsection{Evaluation Setup}
\label{appendix:real_world_eval}
We evaluate DROID-trained policies in both simulation and the real world. For simulation, we use the DROID benchmark~\citep{simevals}, implemented in Isaac Lab~\citep{isaaclab} on top of NVIDIA Isaac Sim~\citep{isaacsim}, and evaluate three pick-and-place tasks: ``put the cube in the bowl,'' ``put the can in the mug,'' and ``put banana in the bin.'' Each rollout lasts 30\,s (450 control steps), and we evaluate replanning horizons of 16 and 24 action steps with 20 trials per task and horizon and results are reported in Table~\ref{tab:replanning_results}. All methods that we train use an action chunk size of 32. For $\pi_{0.5}$, we instead use the official DROID-finetuned checkpoint with its native action chunk size of 15 during simulation evaluation.

For real-world evaluation, we consider three manipulation tasks: placing all three blocks into a bowl,
picking a pen from the cup and placing it onto a table, and folding a towel. We fix the replanning horizon to 16 action steps and conduct 20 trials per task. Each trajectory is terminated after at most 750 control steps, i.e. 50 s of executed actions at $15$\, Hz; time
spent waiting for inference is not counted towards this budget.

\textbf{Place All Blocks in Bowl.}
Three blocks are placed on the table together with a bowl (Figure~\ref{fig:real_world_initial_states}, top left), and the instruction is ``put all the blocks in the bowl.'' A rollout is considered successful if all three blocks are placed in the bowl. We additionally report partial completion rates corresponding to placing at least one, at least two, or all three blocks.

\textbf{Pick Pen from Cup and Place on Table.}
A pen is initially placed inside a cup, and the instruction is ``take the pen from the cup and place it on the table.'' We evaluate four initial configurations that vary the cup placement and the arrangement of distractor objects, including one configuration without distractors (Figure~\ref{fig:real_world_initial_states}, top right), with five rollouts per configuration for a total of 20 trials. A rollout is considered successful if the policy removes the pen from the cup and places it on the table.

\textbf{Fold Towel.}
A towel is initially laid flat on the table, and the instruction is ``fold the towel.'' We evaluate four initial configurations that vary the towel's fold state and orientation: folded in half widthwise, folded widthwise with a rotated orientation, folded in half lengthwise, and fully spread out (Figure~\ref{fig:real_world_initial_states}, bottom). A rollout is considered successful if the robot lifts and folds the towel such that, in total, approximately 40\% or more of the initially exposed towel area is flipped onto the opposite side.

\subsection{Qualitative Rollouts and Failure Modes}
\label{appendix:qualitative_rollouts}

Figures~\ref{fig:rollouts_sim} and~\ref{fig:rollouts_real} illustrate the behavioral differences between \methodname{} and FastWAM across the six DROID evaluation tasks, following the task order in Tables~\ref{tab:replanning_results} and~\ref{tab:real_world_results}. While \methodname{} successfully completes the instructed manipulation tasks, FastWAM either makes partial progress without completing the task or exhibits poor instruction following. These differences appear in both simulation and real-world evaluations: grasping an object does not always lead to a completed placement, and an unsuccessful attempt can be followed by interactions with unrelated objects rather than a retry of the intended manipulation.

\begin{figure}[p]
    \phantomcaption
    \label{fig:rollouts_sim}
    \setlength{\parskip}{0pt}
    \noindent{\footnotesize\bfseries
        Figure~\ref{fig:rollouts_sim}: Simulation rollouts}\par
    \vspace{3pt}\hrule height 0.4pt\vspace{5pt}
    \centering
    \includegraphics[width=0.9\linewidth]{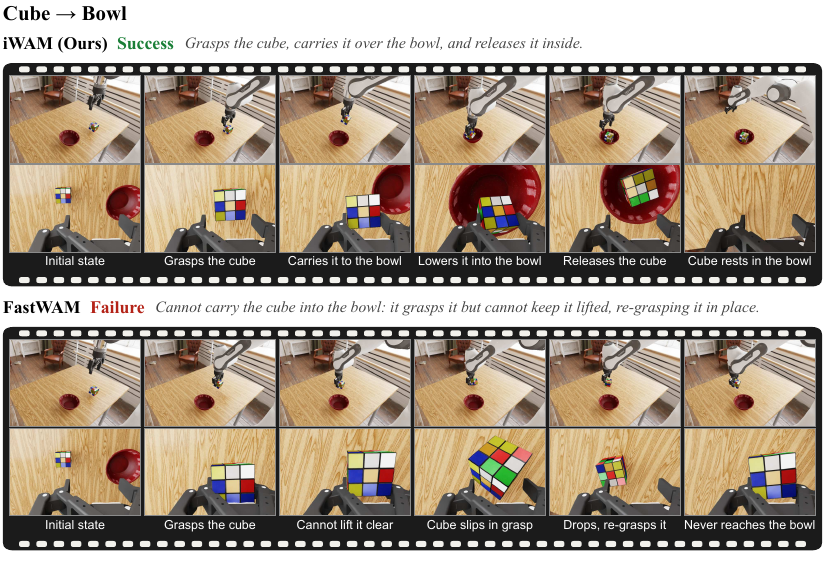}
    \par\vspace{-1pt}
    \includegraphics[width=0.9\linewidth]{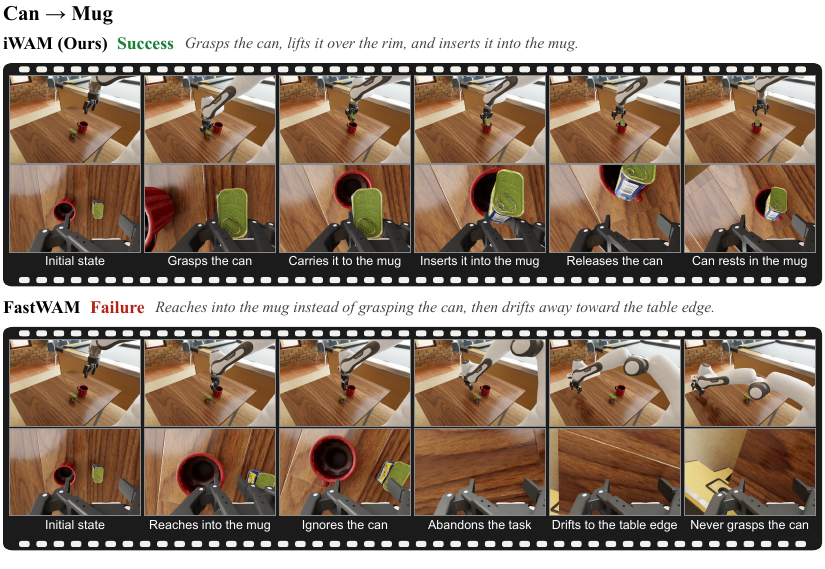}
\end{figure}

\begin{figure}[p]
    \ContinuedFloat
    \setlength{\parskip}{0pt}
    \noindent{\footnotesize\bfseries
        Figure~\ref{fig:rollouts_sim}: Simulation rollouts (continued)}\par
    \vspace{3pt}\hrule height 0.4pt\vspace{5pt}
    \centering
    \includegraphics[width=0.9\linewidth]{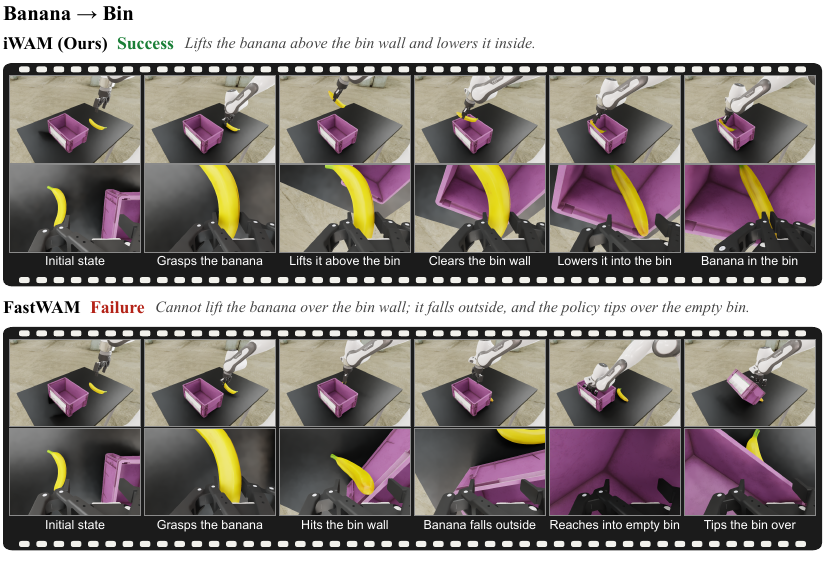}
    \captionsetup{font=footnotesize,labelfont=footnotesize}
    \caption{\textbf{Simulation evaluation behaviors.} \methodname{} maintains its grasp while carrying the cube into the bowl, whereas FastWAM repeatedly loses its grasp. On the can-placement task, \methodname{} lifts the can into the mug, while FastWAM reaches into the mug without grasping the can. On the banana-placement task, \methodname{} lifts the banana high enough to clear the bin wall before lowering it inside; FastWAM fails to clear the wall, drops the banana outside, and subsequently tips over the empty bin.}
\end{figure}

\begin{figure}[p]
    \phantomcaption
    \label{fig:rollouts_real}
    \setlength{\parskip}{0pt}
    \noindent{\footnotesize\bfseries
        Figure~\ref{fig:rollouts_real}: Real-world rollouts}\par
    \vspace{3pt}\hrule height 0.4pt\vspace{5pt}
    \centering
    \includegraphics[width=0.9\linewidth]{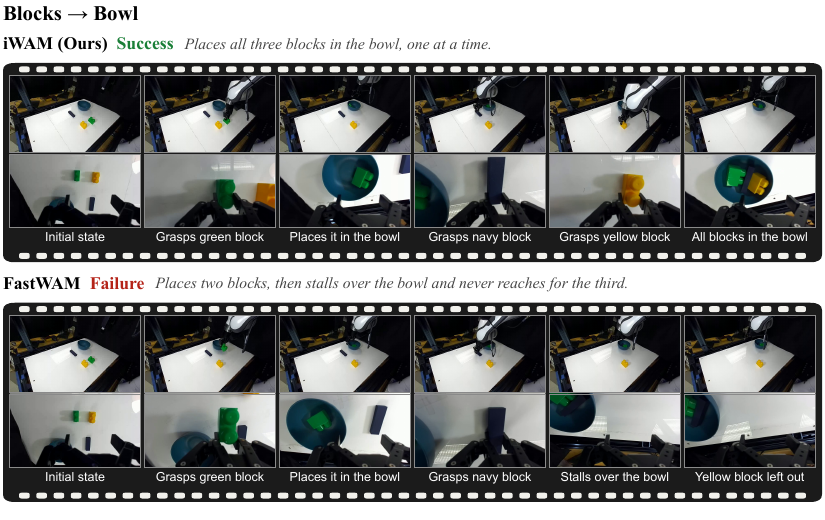}
    \par\vspace{8pt}
    \includegraphics[width=0.9\linewidth]{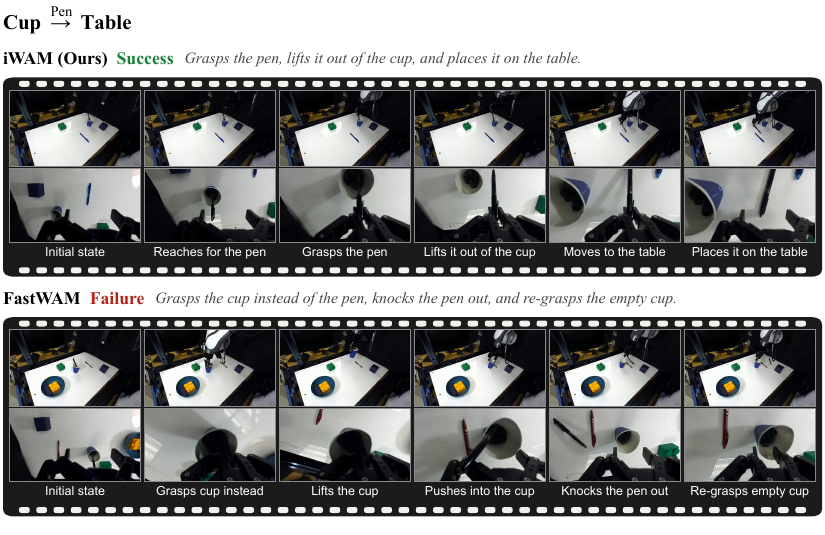}
\end{figure}

\begin{figure}[p]
    \ContinuedFloat
    \setlength{\parskip}{0pt}
    \noindent{\footnotesize\bfseries
        Figure~\ref{fig:rollouts_real}: Real-world rollouts (continued)}\par
    \vspace{3pt}\hrule height 0.4pt\vspace{5pt}
    \centering
    \includegraphics[width=0.9\linewidth]{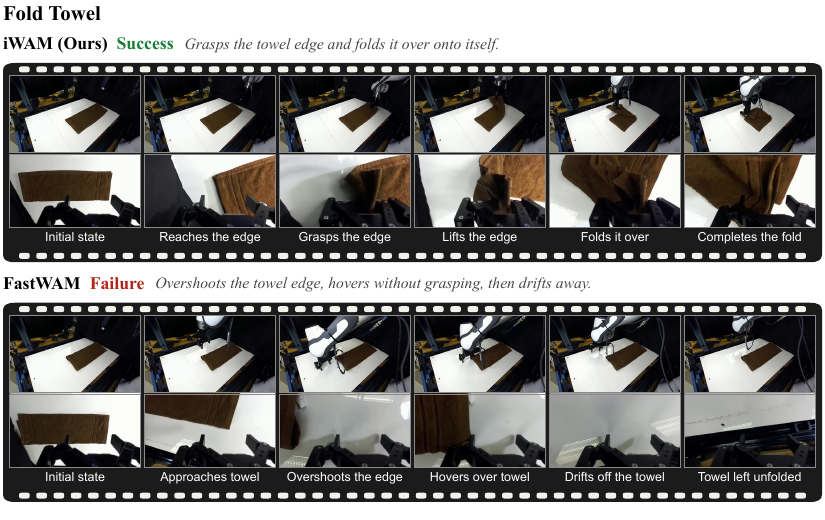}
    \captionsetup{font=footnotesize,labelfont=footnotesize}
    \caption{\textbf{Real-world evaluation behaviors.} \methodname{} places all three blocks in the bowl in succession, whereas FastWAM stops after placing two. On the pen-transfer task, \methodname{} grasps the pen and lifts it out of the cup before placing it on the table; FastWAM instead grasps the cup and knocks the pen out. On the towel-folding task, \methodname{} grasps and lifts the towel edge to fold it over, while FastWAM overshoots the edge and moves away without grasping it.}
\end{figure}

\begin{figure}[!htbp]
    \centering
    \includegraphics[width=0.9\linewidth,keepaspectratio]{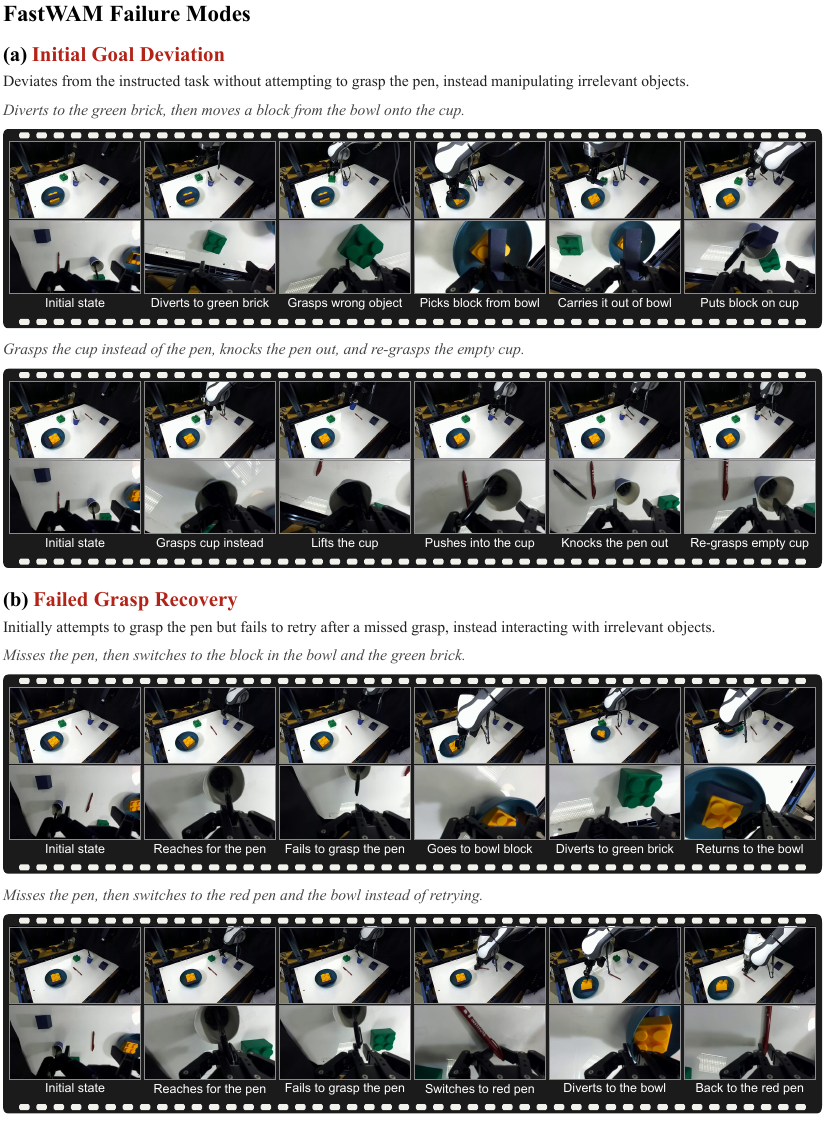}
    \caption{\textbf{FastWAM instruction-following failures on the pen-transfer task.} FastWAM manipulates unrelated objects from the outset (a), or abandons the target object after a missed grasp rather than retrying the intended manipulation (b).}
    \label{fig:fastwam_failure_modes}
\end{figure}

\textbf{FastWAM failure modes.}
Figure~\ref{fig:fastwam_failure_modes} illustrates two forms of instruction-following failure on the pen-transfer task. FastWAM can disregard the instruction from the outset, manipulating unrelated objects without attempting the instructed task. Alternatively, it initially follows the instruction but deviates after a missed grasp, switching to unrelated objects rather than retrying the intended manipulation.

%% file: main.bbl
\begin{thebibliography}{41}
\providecommand{\natexlab}[1]{#1}
\providecommand{\url}[1]{\texttt{#1}}
\expandafter\ifx\csname urlstyle\endcsname\relax
  \providecommand{\doi}[1]{doi: #1}\else
  \providecommand{\doi}{doi: \begingroup \urlstyle{rm}\Url}\fi

\bibitem[Anil et~al.(2022)Anil, Wu, Andreassen, Lewkowycz, Misra, Ramasesh,
  Slone, Gur-Ari, Dyer, and Neyshabur]{cotlength}
Cem Anil, Yuhuai Wu, Anders Andreassen, Aitor Lewkowycz, Vedant Misra, Vinay
  Ramasesh, Ambrose Slone, Guy Gur-Ari, Ethan Dyer, and Behnam Neyshabur.
\newblock Exploring length generalization in large language models.
\newblock \emph{Advances in Neural Information Processing Systems},
  35:\penalty0 38546--38556, 2022.

\bibitem[Bi et~al.(2026)Bi, Tan, Xie, Wang, Huang, Liu, Zhao, Feng, Xiang,
  Rong, et~al.]{motus}
Hongzhe Bi, Hengkai Tan, Shenghao Xie, Zeyuan Wang, Shuhe Huang, Haitian Liu,
  Ruowen Zhao, Yao Feng, Chendong Xiang, Yinze Rong, et~al.
\newblock Motus: A unified latent action world model.
\newblock In \emph{Proceedings of the IEEE/CVF Conference on Computer Vision
  and Pattern Recognition}, pages 35101--35113, 2026.

\bibitem[Boffi et~al.(2026)Boffi, Albergo, and Vanden-Eijnden]{flowmaps}
Nicholas Boffi, Michael Albergo, and Eric Vanden-Eijnden.
\newblock How to build a consistency model: Learning flow maps via
  self-distillation.
\newblock \emph{Advances in Neural Information Processing Systems},
  38:\penalty0 33346--33382, 2026.

\bibitem[Chen et~al.(2025)Chen, Jiang, Zheng, Chen, Su, and Zhu]{gft}
Huayu Chen, Kai Jiang, Kaiwen Zheng, Jianfei Chen, Hang Su, and Jun Zhu.
\newblock Visual generation without guidance.
\newblock \emph{arXiv preprint arXiv:2501.15420}, 2025.

\bibitem[Du et~al.(2023)Du, Yang, Dai, Dai, Nachum, Tenenbaum, Schuurmans, and
  Abbeel]{unipi}
Yilun Du, Sherry Yang, Bo~Dai, Hanjun Dai, Ofir Nachum, Josh Tenenbaum, Dale
  Schuurmans, and Pieter Abbeel.
\newblock Learning universal policies via text-guided video generation.
\newblock \emph{Advances in neural information processing systems},
  36:\penalty0 9156--9172, 2023.

\bibitem[Feng et~al.(2026)Feng, Cui, Ban, and Hsieh]{oneforcing}
Jiaqi Feng, Justin Cui, Yuanhao Ban, and Cho-Jui Hsieh.
\newblock One-forcing: Towards stable one-step autoregressive video generation.
\newblock \emph{arXiv preprint arXiv:2605.23458}, 2026.
\newblock URL \url{https://arxiv.org/abs/2605.23458}.

\bibitem[Frans et~al.(2025)Frans, Hafner, Levine, and Abbeel]{shortcut}
Kevin Frans, Danijar Hafner, Sergey Levine, and Pieter Abbeel.
\newblock One step diffusion via shortcut models.
\newblock In \emph{International Conference on Learning Representations},
  volume 2025, pages 34668--34684, 2025.

\bibitem[Geng et~al.(2026)Geng, Deng, Bai, Kolter, and He]{meanflow}
Zhengyang Geng, Mingyang Deng, Xingjian Bai, Zico Kolter, and Kaiming He.
\newblock Mean flows for one-step generative modeling.
\newblock \emph{Advances in Neural Information Processing Systems},
  38:\penalty0 75460--75482, 2026.

\bibitem[Hou et~al.(2026)Hou, Li, Jia, An, Guo, Leng, Geng, Ze, Harada, Torr,
  Mees, Pollefeys, Liu, Wu, Abbeel, Malik, Du, and Yang]{wamsurvey}
Bohan Hou, Gen Li, Jindou Jia, Tuo An, Xinying Guo, Sicong Leng, Haoran Geng,
  Yanjie Ze, Tatsuya Harada, Philip Torr, Oier Mees, Marc Pollefeys, Zhuang
  Liu, Jiajun Wu, Pieter Abbeel, Jitendra Malik, Yilun Du, and Jianfei Yang.
\newblock World model for robot learning: A comprehensive survey.
\newblock 2026.

\bibitem[Hu et~al.(2025)Hu, Guo, Wang, Chen, Wang, Zhang, Sreenath, Lu, and
  Chen]{vpp}
Yucheng Hu, Yanjiang Guo, Pengchao Wang, Xiaoyu Chen, Yen-Jen Wang, Jianke
  Zhang, Koushil Sreenath, Chaochao Lu, and Jianyu Chen.
\newblock Video prediction policy: A generalist robot policy with predictive
  visual representations.
\newblock In \emph{Proceedings of the 42nd International Conference on Machine
  Learning}, volume 267 of \emph{Proceedings of Machine Learning Research},
  pages 24328--24346. PMLR, 2025.

\bibitem[Huang et~al.(2025)Huang, Wu, Zhou, Miao, and Long]{vid2world}
Siqiao Huang, Jialong Wu, Qixing Zhou, Shangchen Miao, and Mingsheng Long.
\newblock Vid2world: Crafting video diffusion models to interactive world
  models.
\newblock \emph{arXiv preprint arXiv:2505.14357}, 2025.

\bibitem[Jain(2026)]{simevals}
Arhan Jain.
\newblock Droid sim evaluation, 2026.
\newblock URL \url{https://github.com/arhanjain/sim-evals}.

\bibitem[Khazatsky et~al.(2024)Khazatsky, Pertsch, Nair, Balakrishna, Dasari,
  Karamcheti, Nasiriany, Srirama, Chen, Ellis, Fagan, Hejna, Itkina, Lepert,
  Ma, Miller, Wu, Belkhale, Dass, Ha, Jain, Lee, Lee, Memmel, Park,
  Radosavovic, Wang, Zhan, Black, Chi, Hatch, Lin, Lu, Mercat, Rehman, Sanketi,
  Sharma, Simpson, Vuong, Walke, Wulfe, Xiao, Yang, Yavary, Zhao, Agia, Baijal,
  Castro, Chen, Chen, Chung, Drake, Foster, Gao, Guizilini, Herrera, Heo, Hsu,
  Hu, Irshad, Jackson, Le, Li, Lin, Lin, Ma, Maddukuri, Mirchandani, Morton,
  Nguyen, O'Neill, Scalise, Seale, Son, Tian, Tran, Wang, Wu, Xie, Yang, Yin,
  Zhang, Bastani, Berseth, Bohg, Goldberg, Gupta, Gupta, Jayaraman, Lim, Malik,
  Martín-Martín, Ramamoorthy, Sadigh, Song, Wu, Yip, Zhu, Kollar, Levine, and
  Finn]{droid}
Alexander Khazatsky, Karl Pertsch, Suraj Nair, Ashwin Balakrishna, Sudeep
  Dasari, Siddharth Karamcheti, Soroush Nasiriany, Mohan~Kumar Srirama,
  Lawrence~Yunliang Chen, Kirsty Ellis, Peter~David Fagan, Joey Hejna, Masha
  Itkina, Marion Lepert, Yecheng~Jason Ma, Patrick~Tree Miller, Jimmy Wu,
  Suneel Belkhale, Shivin Dass, Huy Ha, Arhan Jain, Abraham Lee, Youngwoon Lee,
  Marius Memmel, Sungjae Park, Ilija Radosavovic, Kaiyuan Wang, Albert Zhan,
  Kevin Black, Cheng Chi, Kyle~Beltran Hatch, Shan Lin, Jingpei Lu, Jean
  Mercat, Abdul Rehman, Pannag~R Sanketi, Archit Sharma, Cody Simpson, Quan
  Vuong, Homer~Rich Walke, Blake Wulfe, Ted Xiao, Jonathan~Heewon Yang, Arefeh
  Yavary, Tony~Z. Zhao, Christopher Agia, Rohan Baijal, Mateo~Guaman Castro,
  Daphne Chen, Qiuyu Chen, Trinity Chung, Jaimyn Drake, Ethan~Paul Foster,
  Jensen Gao, Vitor Guizilini, David~Antonio Herrera, Minho Heo, Kyle Hsu,
  Jiaheng Hu, Muhammad~Zubair Irshad, Donovon Jackson, Charlotte Le, Yunshuang
  Li, Kevin Lin, Roy Lin, Zehan Ma, Abhiram Maddukuri, Suvir Mirchandani,
  Daniel Morton, Tony Nguyen, Abigail O'Neill, Rosario Scalise, Derick Seale,
  Victor Son, Stephen Tian, Emi Tran, Andrew~E. Wang, Yilin Wu, Annie Xie,
  Jingyun Yang, Patrick Yin, Yunchu Zhang, Osbert Bastani, Glen Berseth,
  Jeannette Bohg, Ken Goldberg, Abhinav Gupta, Abhishek Gupta, Dinesh
  Jayaraman, Joseph~J Lim, Jitendra Malik, Roberto Martín-Martín, Subramanian
  Ramamoorthy, Dorsa Sadigh, Shuran Song, Jiajun Wu, Michael~C. Yip, Yuke Zhu,
  Thomas Kollar, Sergey Levine, and Chelsea Finn.
\newblock Droid: A large-scale in-the-wild robot manipulation dataset.
\newblock 2024.

\bibitem[Kim et~al.(2026)Kim, Gao, Lin, Lin, Ge, Lam, Liang, Song, Liu, Finn,
  and Gu]{cosmospolicy}
Moo~Jin Kim, Yihuai Gao, Tsung-Yi Lin, Yen-Chen Lin, Yunhao Ge, Grace Lam,
  Percy Liang, Shuran Song, Ming-Yu Liu, Chelsea Finn, and Jinwei Gu.
\newblock Cosmos policy: Fine-tuning video models for visuomotor control and
  planning.
\newblock In \emph{The Fourteenth International Conference on Learning
  Representations}, 2026.

\bibitem[Li et~al.(2026{\natexlab{a}})Li, Zeng, Lu, Zhu, Ouyang, Wang, Cheng,
  Shen, and Zhang]{aad1}
Haobo Li, Yanhong Zeng, Yunhong Lu, Jiapeng Zhu, Hao Ouyang, Qiuyu Wang,
  Ka~Leong Cheng, Yujun Shen, and Zhipeng Zhang.
\newblock Aad-1: Asymmetric adversarial distillation for one-step
  autoregressive video generation.
\newblock \emph{arXiv preprint arXiv:2606.03972}, 2026{\natexlab{a}}.

\bibitem[Li and He(2024)]{bcm}
Liangchen Li and Jiajun He.
\newblock Bidirectional consistency models.
\newblock \emph{arXiv preprint arXiv:2403.18035}, 2024.

\bibitem[Li et~al.(2026{\natexlab{b}})Li, Zhang, Luo, Yang, Wang, Han, Yu, Gao,
  Xue, Zhu, et~al.]{causalWM}
Lin Li, Qihang Zhang, Yiming Luo, Shuai Yang, Ruilin Wang, Fei Han, Mingrui Yu,
  Zelin Gao, Nan Xue, Xing Zhu, et~al.
\newblock Causal world modeling for robot control.
\newblock \emph{arXiv preprint arXiv:2601.21998}, 2026{\natexlab{b}}.

\bibitem[Li et~al.()Li, Gao, Sadigh, and Song]{uva}
Shuang Li, Yihuai Gao, Dorsa Sadigh, and Shuran Song.
\newblock Unified video action model, 2025.
\newblock \emph{arXiv preprint arXiv:2503.00200}.

\bibitem[Liang et~al.(2024)Liang, Liu, Ozguroglu, Sudhakar, Dave, Tokmakov,
  Song, and Vondrick]{dreamitate}
Junbang Liang, Ruoshi Liu, Ege Ozguroglu, Sruthi Sudhakar, Achal Dave, Pavel
  Tokmakov, Shuran Song, and Carl Vondrick.
\newblock Dreamitate: Real-world visuomotor policy learning via video
  generation, 2024.

\bibitem[Liang et~al.(2025)Liang, Tokmakov, Liu, Sudhakar, Shah, Ambrus, and
  Vondrick]{disjointvideo}
Junbang Liang, Pavel Tokmakov, Ruoshi Liu, Sruthi Sudhakar, Paarth Shah, Rares
  Ambrus, and Carl Vondrick.
\newblock Video generators are robot policies.
\newblock \emph{arXiv preprint arXiv:2508.00795}, 2025.

\bibitem[Lipman et~al.(2022)Lipman, Chen, Ben-Hamu, Nickel, and Le]{originflow}
Yaron Lipman, Ricky~TQ Chen, Heli Ben-Hamu, Maximilian Nickel, and Matt Le.
\newblock Flow matching for generative modeling.
\newblock \emph{arXiv preprint arXiv:2210.02747}, 2022.

\bibitem[Liu et~al.(2023)Liu, Zhu, Gao, Feng, Liu, Zhu, and Stone]{libero}
Bo~Liu, Yifeng Zhu, Chongkai Gao, Yihao Feng, Qiang Liu, Yuke Zhu, and Peter
  Stone.
\newblock Libero: Benchmarking knowledge transfer for lifelong robot learning.
\newblock \emph{arXiv preprint arXiv:2306.03310}, 2023.

\bibitem[Mirzadeh et~al.(2025)Mirzadeh, Alizadeh-Vahid, Shahrokhi, Tuzel,
  Bengio, and Farajtabar]{gsmperturb}
Iman Mirzadeh, Keivan Alizadeh-Vahid, Hooman Shahrokhi, Oncel Tuzel, Samy
  Bengio, and Mehrdad Farajtabar.
\newblock Gsm-symbolic: Understanding the limitations of mathematical reasoning
  in large language models.
\newblock In \emph{International Conference on Learning Representations},
  volume 2025, pages 94743--94765, 2025.

\bibitem[Mittal et~al.(2025)Mittal, Roth, Tigue, Richard, Zhang, Du,
  Serrano-Munoz, Yao, Zurbr{\"u}gg, Rudin, et~al.]{isaaclab}
Mayank Mittal, Pascal Roth, James Tigue, Antoine Richard, Octi Zhang, Peter Du,
  Antonio Serrano-Munoz, Xinjie Yao, Ren{\'e} Zurbr{\"u}gg, Nikita Rudin,
  et~al.
\newblock Isaac lab: A gpu-accelerated simulation framework for multi-modal
  robot learning.
\newblock \emph{arXiv preprint arXiv:2511.04831}, 2025.

\bibitem[Ni et~al.(2026)Ni, Zhou, Zhou, Liu, Wang, Zhu, Wang, Deng, Ye, Li,
  et~al.]{gigaworld}
Chaojun Ni, Xinyu Zhou, Yukun Zhou, Jingyu Liu, Xiaofeng Wang, Zheng Zhu, Yang
  Wang, Qiuping Deng, Yun Ye, Hao Li, et~al.
\newblock Gigaworld-policy: An efficient action-centered world--action model.
\newblock In \emph{European Conference on Computer Vision}, pages 315--333.
  Springer, 2026.

\bibitem[{NVIDIA}()]{isaacsim}
{NVIDIA}.
\newblock Isaac sim.
\newblock URL \url{https://github.com/isaac-sim/IsaacSim}.

\bibitem[Pan et~al.(2025)Pan, Anantharaman, Huang, Jin, Pfrommer, Yuan,
  Permenter, Qu, Boffi, Shi, and Simchowitz]{mip}
Chaoyi Pan, Giri Anantharaman, Nai-Chieh Huang, Claire Jin, Daniel Pfrommer,
  Chenyang Yuan, Frank Permenter, Guannan Qu, Nicholas Boffi, Guanya Shi, and
  Max Simchowitz.
\newblock Much ado about noising: Dispelling the myths of generative robotic
  control.
\newblock 2025.
\newblock URL \url{https://arxiv.org/abs/2512.01809}.

\bibitem[Salimans and Ho(2022)]{progressive}
Tim Salimans and Jonathan Ho.
\newblock Progressive distillation for fast sampling of diffusion models.
\newblock \emph{arXiv preprint arXiv:2202.00512}, 2022.

\bibitem[Setlur et~al.(2025)Setlur, Qu, Yang, Zhang, Smith, and
  Kumar]{setlur2025opt}
Amrith Setlur, Yuxiao Qu, Matthew Yang, Lunjun Zhang, Virginia Smith, and
  Aviral Kumar.
\newblock Optimizing llm test-time compute involves solving a meta-rl problem.
\newblock
  \url{https://blog.ml.cmu.edu/2025/01/08/optimizing-llm-test-time-compute-involves-solving-a-meta-rl-problem/},
  2025.
\newblock CMU MLD Blog.

\bibitem[Snell et~al.(2024)Snell, Lee, Xu, and Kumar]{snell2024scaling}
Charlie Snell, Jaehoon Lee, Kelvin Xu, and Aviral Kumar.
\newblock Scaling llm test-time compute optimally can be more effective than
  scaling model parameters.
\newblock \emph{arXiv preprint arXiv:2408.03314}, 2024.

\bibitem[Song et~al.(2021)Song, Meng, and Ermon]{ddim}
Jiaming Song, Chenlin Meng, and Stefano Ermon.
\newblock Denoising diffusion implicit models.
\newblock In \emph{International Conference on Learning Representations}, 2021.

\bibitem[Song et~al.(2023)Song, Dhariwal, Chen, and Sutskever]{consistency}
Yang Song, Prafulla Dhariwal, Mark Chen, and Ilya Sutskever.
\newblock Consistency models.
\newblock In \emph{Proceedings of the 40th International Conference on Machine
  Learning}, volume 202 of \emph{Proceedings of Machine Learning Research},
  pages 32211--32252. PMLR, 2023.

\bibitem[{Team Wan} et~al.(2025){Team Wan}, Wang, Ai, Wen, Mao,
  et~al.]{wan2025}
{Team Wan}, Ang Wang, Baole Ai, Bin Wen, Chaojie Mao, et~al.
\newblock Wan: Open and advanced large-scale video generative models.
\newblock \emph{arXiv preprint arXiv:2503.20314}, 2025.

\bibitem[Wen et~al.(2024)Wen, Lin, Zhu, Han, Xu, Zhao, and Liang]{vidman}
Youpeng Wen, Junfan Lin, Yi~Zhu, Jianhua Han, Hang Xu, Shen Zhao, and Xiaodan
  Liang.
\newblock Vidman: Exploiting implicit dynamics from video diffusion model for
  effective robot manipulation.
\newblock \emph{Advances in Neural Information Processing Systems},
  37:\penalty0 41051--41075, 2024.

\bibitem[Ye et~al.(2026)Ye, Ge, Zheng, Gao, Yu, Kurian, Indupuru, Tan, Zhu,
  Xiang, Malik, Lee, Liang, Ranawaka, Gu, Xu, Wang, Hu, Narayan, Bjorck, Wang,
  Kim, Niu, Zheng, Xie, Wu, Wang, Julian, Xu, Du, Chebotar, Reed, Kautz, Zhu,
  Fan, and Jang]{dreamzero}
Seonghyeon Ye, Yunhao Ge, Kaiyuan Zheng, Shenyuan Gao, Sihyun Yu, George
  Kurian, Suneel Indupuru, You~Liang Tan, Chuning Zhu, Jiannan Xiang, Ayaan
  Malik, Kyungmin Lee, William Liang, Nadun Ranawaka, Jiasheng Gu, Yinzhen Xu,
  Guanzhi Wang, Fengyuan Hu, Avnish Narayan, Johan Bjorck, Jing Wang, Gwanghyun
  Kim, Dantong Niu, Ruijie Zheng, Yuqi Xie, Jimmy Wu, Qi~Wang, Ryan Julian,
  Danfei Xu, Yilun Du, Yevgen Chebotar, Scott Reed, Jan Kautz, Yuke Zhu,
  Linxi~"Jim" Fan, and Joel Jang.
\newblock World action models are zero-shot policies, 2026.
\newblock URL \url{https://arxiv.org/abs/2602.15922}.

\bibitem[Yuan et~al.(2026)Yuan, Dong, Liu, and Zhao]{yuan2026fastwam}
Tianyuan Yuan, Zibin Dong, Yicheng Liu, and Hang Zhao.
\newblock Fast-wam: Do world action models need test-time future imagination?
\newblock \emph{arXiv preprint arXiv:2603.16666}, 2026.
\newblock URL \url{https://arxiv.org/abs/2603.16666}.

\bibitem[Zheng et~al.(2026{\natexlab{a}})Zheng, He, Zhao, Zhang, Chen, Chen,
  Lin, Liu, Zhu, and Ma]{crcm}
Kaiwen Zheng, Guande He, Min Zhao, Jintao Zhang, Huayu Chen, Jianfei Chen,
  Chen-Hsuan Lin, Ming-Yu Liu, Jun Zhu, and Qianli Ma.
\newblock Causal-rcm: A unified teacher-forcing and self-forcing open recipe
  for autoregressive diffusion distillation in streaming video generation and
  interactive world models.
\newblock \emph{arXiv preprint arXiv:2606.25473}, 2026{\natexlab{a}}.

\bibitem[Zheng et~al.(2026{\natexlab{b}})Zheng, Wang, Ma, Chen, Zhang, Balaji,
  Chen, Liu, Zhu, and Zhang]{rcm}
Kaiwen Zheng, Yuji Wang, Qianli Ma, Huayu Chen, Jintao Zhang, Yogesh Balaji,
  Jianfei Chen, Ming-Yu Liu, Jun Zhu, and Qinsheng Zhang.
\newblock Large scale diffusion distillation via score-regularized
  continuous-time consistency.
\newblock In \emph{International Conference on Learning Representations},
  volume 2026, pages 2582--2603, 2026{\natexlab{b}}.

\bibitem[Zhou et~al.(2026)Zhou, Parger, Haque, and Song]{tvm}
Linqi Zhou, Mathias Parger, Ayaan Haque, and Jiaming Song.
\newblock Terminal velocity matching.
\newblock In \emph{International Conference on Learning Representations},
  volume 2026, pages 87506--87530, 2026.

\bibitem[Zhou et~al.(2025)Zhou, Xu, Tie, Chen, Zhang, Chu, Zhou, and
  Sun]{liberopro}
Xueyang Zhou, Yangming Xu, Guiyao Tie, Yongchao Chen, Guowen Zhang, Duanfeng
  Chu, Pan Zhou, and Lichao Sun.
\newblock Libero-pro: Towards robust and fair evaluation of
  vision-language-action models beyond memorization.
\newblock \emph{arXiv preprint arXiv:2510.03827}, 2025.

\bibitem[Zhu et~al.(2025)Zhu, Yu, Feng, Burchfiel, Shah, and Gupta]{unifiedWM}
Chuning Zhu, Raymond Yu, Siyuan Feng, Benjamin Burchfiel, Paarth Shah, and
  Abhishek Gupta.
\newblock Unified world models: Coupling video and action diffusion for
  pretraining on large robotic datasets.
\newblock \emph{arXiv preprint arXiv:2504.02792}, 2025.

\end{thebibliography}
